\documentclass[10pt,journal,compsoc]{IEEEtran}

\ifCLASSOPTIONcompsoc
  \usepackage[nocompress]{cite}
\else
  \usepackage{cite}
\fi

\ifCLASSINFOpdf
  \usepackage[pdftex]{graphicx}
\else
  \usepackage[dvips]{graphicx}
\fi

\usepackage{amsmath}
\usepackage{array}
\usepackage{algorithm}
\usepackage[noend]{algpseudocode}
\makeatletter
\def\BState{\State\hskip-\ALG@thistlm}
\makeatother

\ifCLASSOPTIONcompsoc
  \usepackage[caption=false,font=footnotesize,labelfont=sf,textfont=sf]{subfig}
\else
  \usepackage[caption=false,font=footnotesize]{subfig}
\fi

\usepackage{fixltx2e}
\usepackage{stfloats}
\usepackage{multicol}
\usepackage{comment}
\usepackage{lscape}
\usepackage{longtable}

\ifCLASSOPTIONcaptionsoff
  \usepackage[nomarkers]{endfloat}
 \let\MYoriglatexcaption\caption
 \renewcommand{\caption}[2][\relax]{\MYoriglatexcaption[#2]{#2}}
\fi

\usepackage{url}

\makeatletter
\newcommand*\titleheader[1]{\gdef\@titleheader{#1}}
\AtBeginDocument{%
	\let\st@red@title\@title%
	\def\@title{%
		\bgroup\normalfont\small\centering\@titleheader\par\egroup
		\vskip0.0em\st@red@title}
}
\makeatother

\title{SECRET: Semantically Enhanced Classification of Real-world Tasks}

\titleheader{\textsuperscript{\textcopyright} 2020 IEEE.  Personal use of this material is permitted. Permission from IEEE must  be obtained for all other uses, in any current or future media, including reprinting/republishing this material for advertising or promotional purposes, creating new collective works, for resale or redistribution to servers or lists, or reuse of any copyrighted component of this work in other works.}%

\begin{document}


\author{Ayten Ozge~Akmandor,~\IEEEmembership{Student Member,~IEEE,}
~Jorge~Ortiz,~\IEEEmembership{Member,~IEEE,}~Irene~Manotas,~Bongjun~Ko,
	and~Niraj K.~Jha,~\IEEEmembership{Fellow,~IEEE}
\thanks{This work was supported by an IBM summer internship and NSF Grant 
No. CNS-1617640. Ayten Ozge Akmandor and Niraj K. Jha are with the 
Department of Electrical Engineering, Princeton University, Princeton,
NJ, 08544, USA, e-mail:\{akmandor, jha\}@princeton.edu. Jorge Ortiz is with the 
Department of Electrical and Computer Engineering, Rutgers University, Piscataway, 
NJ, 08854, USA, e-mail: jorge.ortiz@rutgers.edu. Irene Manotas and Bongjun Ko are 
with IBM Thomas J. Watson Research Center, Yorktown Heights, NY, 10598 USA, 
e-mail:\{Irene.Manotas, bongjun\_ko\}@us.ibm.com.}}


\IEEEtitleabstractindextext{%
\begin{abstract}
Supervised machine learning (ML) algorithms are aimed at maximizing 
classification performance under available energy and storage constraints. 
They try to map the training data to the corresponding labels while ensuring 
generalizability to unseen data.  However, they do not integrate meaning-based 
relationships among labels in the decision process.  On the other hand, natural language 
processing (NLP) algorithms emphasize the importance of semantic information. 
In this paper, we synthesize the complementary
advantages of supervised ML and NLP
algorithms into one method that we refer to as 
SECRET (Semantically Enhanced Classification of REal-world Tasks). SECRET 
performs classifications by fusing the semantic information of the
labels with the available data: it combines the feature space of the 
supervised algorithms with the semantic space of the NLP algorithms
and predicts labels based on this joint space. Experimental results
indicate that, compared to traditional supervised learning, SECRET achieves 
up to 14.0\% accuracy and 13.1\% F1 score improvements. 
Moreover, compared to ensemble methods, SECRET achieves up to 12.7\% accuracy and 
13.3\% F1 score improvements.  This points to a new research direction for supervised 
classification based on incorporation of semantic information. 
\end{abstract}

}

\maketitle

\IEEEdisplaynontitleabstractindextext

\IEEEpeerreviewmaketitle

\IEEEraisesectionheading{\section{Introduction}\label{sec:introduction}}

\IEEEPARstart{S}{ignificant} progress has been made in natural language 
processing (NLP) and supervised machine learning (ML) algorithms over the past 
two decades. NLP successes include machine translation, speech/emotion/sentiment 
recognition, machine reading, and social media mining 
\cite{hirschberg2015advances}. Hence, NLP is beginning to become widely used 
in real-world applications that include either speech or text. Supervised 
ML algorithms excel at modeling the data-label relationship 
while maximizing performance and minimizing energy consumption and latency. 

Supervised ML algorithms train on feature-label pairs 
to model the application of interest and predict labels. The label involves 
semantic information. Palatucci et al.~\cite{palatucci2009zero} use this 
information through vector representations of words to find the novel class 
within the dataset. Karpathy and Fei-Fei \cite{karpathy2015deep} generate 
figure captions based on the collective use of image datasets and word 
embeddings. Such studies indicate that data features and semantic 
relationships correlate well. However, current supervised ML
algorithms do not utilize such correlations in the decision-making
(prediction) process. Their decisions are only based on the feature-label 
relationship, while neglecting significant information hidden in the 
labels, i.e., meaning-based (semantic) relationships among labels. 
Thus, they are not able to exploit synergies between the feature and
semantic spaces.

In this article, we show the above synergies can be exploited to improve the prediction performance of ML algorithms. Our method, called SECRET, combines vector representations of labels in the semantic space with available data in the feature space within various operations (e.g., ML hyperparameter optimization and confidence score computation) to make the final decisions (assign labels to datapoints). Since SECRET does not target any particular ML algorithm or data
structure, it is widely applicable.

The main contributions of this article are as follows:

\begin{itemize}
	\item We introduce a dual-space ML decision process called 
SECRET. It combines the new dimension (semantic space) with the traditional 
(single-space) classifiers that operate in the feature space.
Thus, SECRET not only utilizes available data-label pairs, but also
takes advantage of meaning-based (semantic) relationships among labels to 
perform classification for a given real-world task. 
	\item We demonstrate the general applicability of SECRET on various
supervised ML algorithms and a wide range of datasets for various real-world tasks. 
	\item We demonstrate the advantages of SECRET's new dimension
(semantic space) through detailed comparisons with traditional ML approaches 
that have the same processing and information (except semantic) resources.
	\item We compare the semantic space ML model with traditional approaches. We shed light on 
how SECRET builds the semantic space component and its impact on overall classification performance.
\end{itemize}

The remainder of the article is organized as follows. Section \ref{background} provides background 
information on Bayesian optimization and semantic vector representation
of words. Section \ref{motivation} 
provides the motivation behind SECRET's dual-space ML decision process.
 Section \ref{methodology} introduces the methodologies
underpinning the SECRET architecture, data processing, hyperparameter tuning, 
ML algorithm training in the feature and semantic spaces, confidence
score calculation, and decision process. Section \ref{results} presents
experimental results and provides comparisons with traditional ML
approaches. Section \ref{related_work} presents related work from the literature and points 
out the novelty of SECRET. Section \ref{discussion} provides a detailed discussion on SECRET from different perspectives. Finally, Section \ref{conclusion} discusses future research
directions and concludes the article.

\section{Background} \label{background}
In this section, we discuss background material that will help with understanding of the rest
of the article.  We discuss Bayesian optimization for hyperparameter tuning and semantic vector 
representation of words.

\subsection{Bayesian Optimization for Hyperparameter Tuning} \label{hyperparameter_tuning}
The selected set of hyperparameter values has a direct impact on classification/regression 
performance. Hand-tuning, random search, grid search, and Bayesian optimization are commonly used 
methods for finding the best set of hyperparameter values. 
In this work, we adopt Bayesian optimization as it is known, in general, to provide an unbiased analysis and higher 
classification/regression performance, while requiring a small number of iterations due 
to the utilization of results from past iterations \cite{bergstra2013making}.

Bayesian optimization integrates exploration and exploitation. It starts with a prior belief over 
the unknown objective function. It then evaluates the optimization goal function with available 
data (target hyperparameter values chosen for the iteration). Based on input data and the 
corresponding optimization goal outputs, it updates the beliefs and selects the next set of 
hyperparameter values to be evaluated. The process is repeated until a maximum number of iterations 
is reached \cite{shahriari2016taking}. 

\subsection{Semantic Vector Models of Words} \label{semantic_vector_models}
Semantic vector models assign a compact real-valued vector to each word in a dictionary. The vector 
captures the word's semantic relationships with the remaining words in the dictionary. Words 
with close meanings are represented by closely-spaced vectors in the semantic space. Some of
the algorithms that derive semantic vector word representations are Skip-gram and 
Continuous Bag-of-Words (CBOW) architectures of word2vec \cite{mikolov2013efficient}, 
GloVe \cite{pennington2014glove}, vLBL \cite{mnih2013learning}, ivLBL \cite{mnih2013learning}, 
Hellinger PCA \cite{lebret2013word}, and recurrent neural networks \cite{mikolov2010recurrent}.

GloVe is an unsupervised method. It uses the co-occurrence ratio of words within a pre-specified 
window length to obtain the word vectors. Use of this ratio enhances the distinction 
between two relevant words or a relevant word and an irrelevant one. The GloVe algorithm is based on 
weighted least squares regression. As shown in Eq.~\ref{glove_main}, it aims to minimize the 
difference between the scalar product of the two word vectors and the logarithm of their co-occurrence 
value. Weights are used to avoid dominance (overweighting) by both very frequent and rare 
co-occurrences. The corresponding weighting function is shown in Eq.~\ref{glove_weight}. In 
\cite{pennington2014glove}, $\alpha = \frac{3}{4}$ has been found to yield good results. 

\begin{equation}
\begin{aligned}
J = \sum_{i=1}^{V} \sum_{j=1}^{V} f(X_{ij})(w_{i}^T*w_{j}+b_{i}+b_{j}-log(X_{ij}))^2, \\
\text{where } V \text{: vocabulary size}, \\
f(x) \text{: weighting function}, \\
X \text{: co-occurrence matrix}, \\
w \text{: word vector}, \\
b \text{: bias}.
\end{aligned}
\label{glove_main}
\end{equation}

\begin{equation}
\begin{aligned}
f(x) =
\begin{cases}
(x/x_{max})^\alpha,& \text{if } x<x_{max}\\
1,              & \text{otherwise}
\end{cases}
\end{aligned}
\label{glove_weight}
\end{equation}

\section{Motivation} \label{motivation}

\begin{figure}[t!]
	\centering
	\subfloat[]{\includegraphics[width=3.1in]{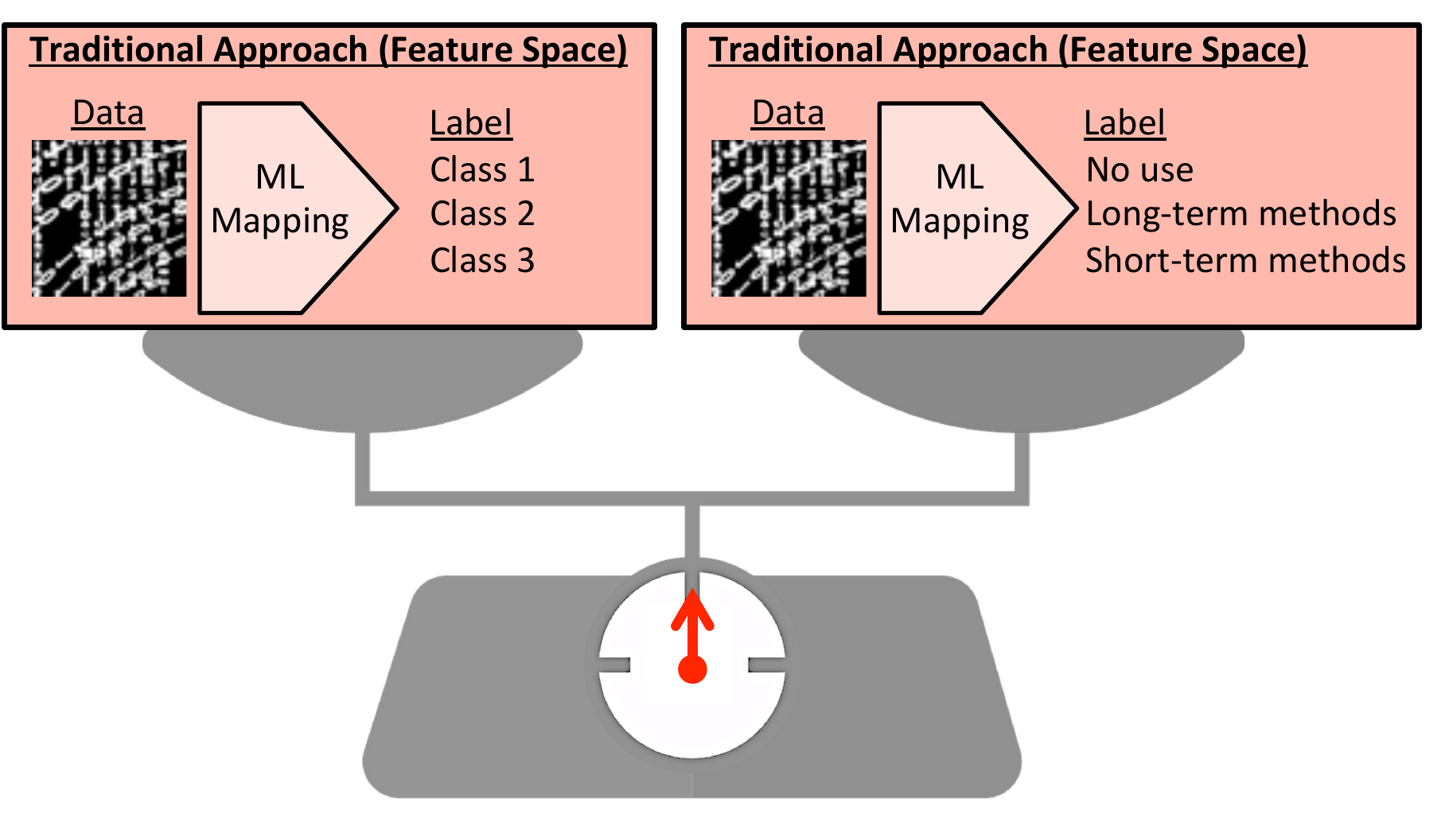}%
		\label{traditional_approach}}
	\vspace{-0.4cm}
	\hfil
	\subfloat[]{\includegraphics[width=3.1in]{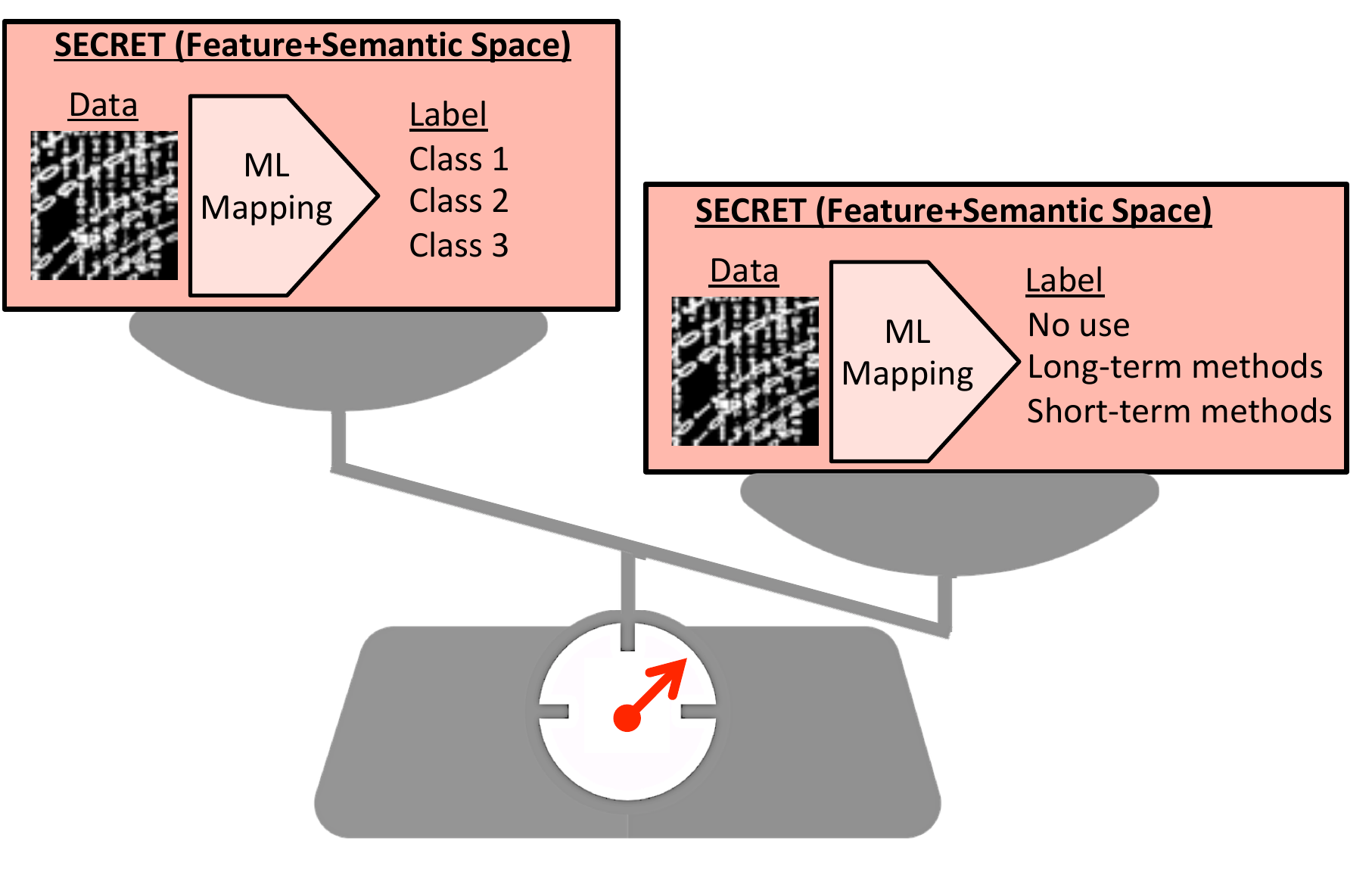}%
		\label{secret_approach}}
	\vspace{-0.3cm}
	\hfil

	\caption{ML data-label mapping responses of (a) the traditional approach and (b) SECRET. The scale signifies whether the data-label models are equal (equilibrium state) for the two types of labels. }
	\label{comparison}
	
\end{figure}

Various features (characteristics) can be extracted from data and correlated with 
corresponding circumstances (labels) of interest in supervised ML. 
The features constitute the feature space.  As an example, in healthcare, 
the labels may be disease names, therapy methods, or health states, whereas 
in the chemical industry, the labels may be chemical names, model simulation 
states, or stability test results.  Although labels differ from one
application to another, they all lead to some action being taken
based on the assigned label. The action can be reporting an anomaly,
continuing the process, switching states, scaling parameters,
etc. Since the assigned labels impact future actions, they need to
be interpretable by either humans or machines. This means that the
labels also carry semantic information. However, current supervised
classifiers do not take advantage of this semantic information.
Consider a dataset that has `long-term methods,' `short-term methods,' and 
`no use' as labels. As depicted in Fig.~\ref{comparison}, current 
supervised ML algorithms will result in the same data-label model even if 
we replace the labels with `class 1,' `class 2,' and `class 3.' However, 
`long-term methods' and `short-term methods' are semantically more similar but less similar to `no use,' as indicated by the squared Euclidean distances in Fig. \ref{cmc_label_position_ss}. It would be advantageous to exploit this semantic 
relationship during classification.

\begin{figure}[t!]
	\centering
	\subfloat[]{\includegraphics[width=2.8in]{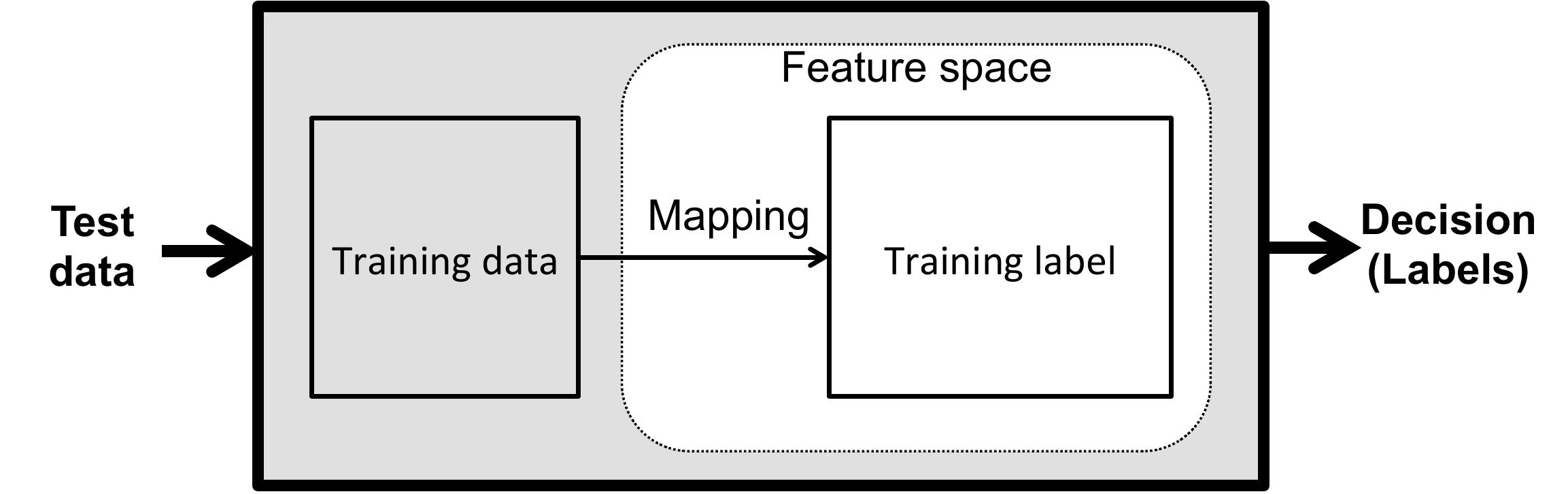}%
		\label{fs_classifier}}
	\vspace{-0.2cm}
	\hfil
	\subfloat[]{\includegraphics[width=3.3in]{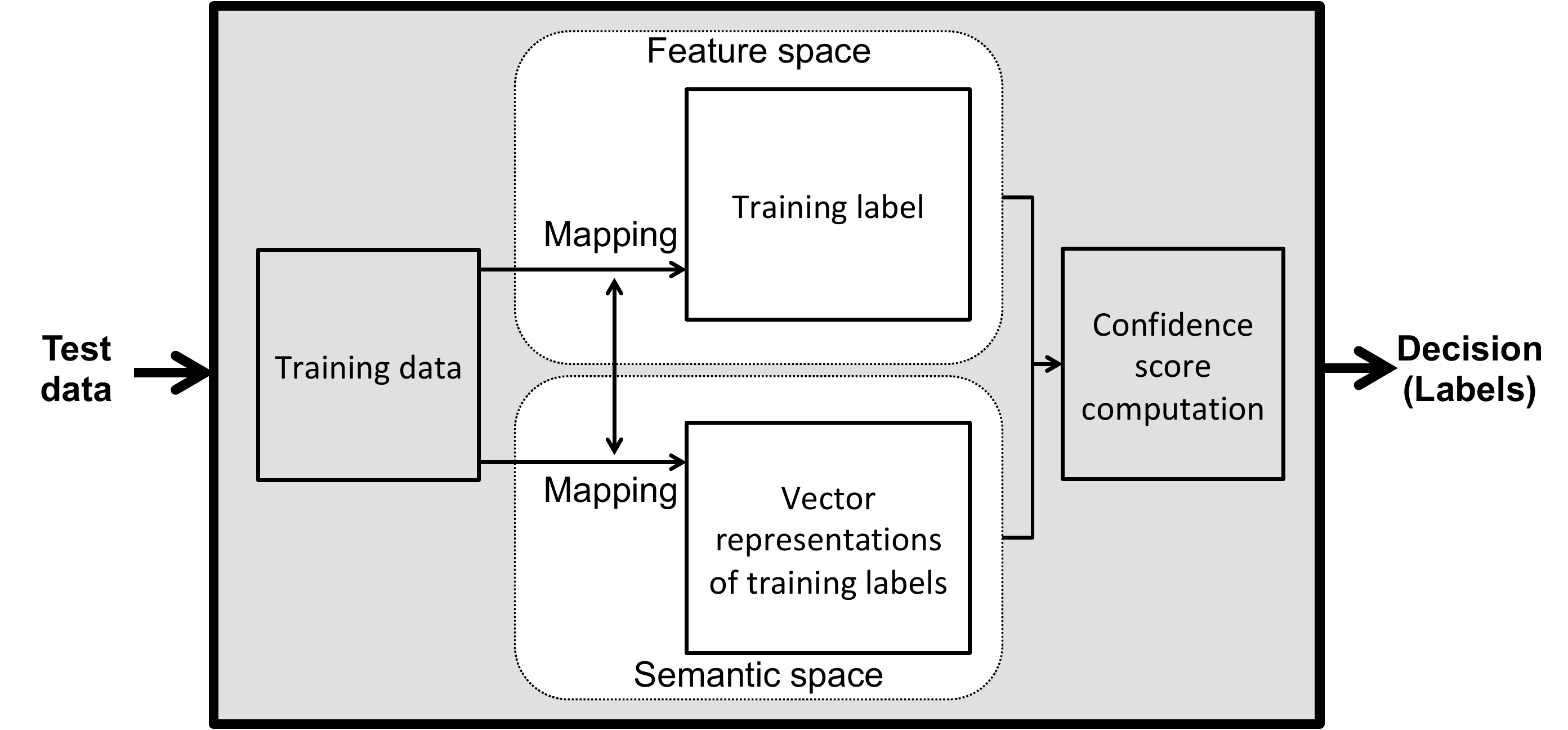}%
		\label{secret_classifier}}
	
	\hfil
	\vspace{-0.4cm}
	\caption{Classification methods: (a) traditional and (b) SECRET.}
	\label{fs_secret_classifier}
	\vspace{-0.6cm}
\end{figure}

SECRET addresses the above problem through a dual-space
classification approach. As shown in Fig.~\ref{fs_classifier}, traditional supervised learning operates in the feature space. SECRET, on the other hand, also incorporates class affinity and dissimilarity information into the decision process, as shown by the `Semantic space' block in Fig.~\ref{secret_classifier}. This property enables SECRET to make informed decisions on class labels. SECRET makes the final decision (labeling) by integrating information from both the feature and semantic spaces. Therefore, it is able to deliver higher classification performance relative to traditional approaches because of its reliance on a richer 
semantic+feature space. Although this example only exploits the semantic+feature space, there may be other as-yet-undiscovered spaces that could also be integrated into SECRET in a similar fashion. 

\section{Methodology} \label{methodology}
In this section, we describe SECRET's data processing and dual-space classification procedure 
in detail.

\subsection{The SECRET Architecture} \label{secret_arch}
SECRET integrates information from two sources: feature space and semantic space. 

\begin{itemize}
	\item The feature space includes data, extracted features (if available), and the 
corresponding labels.
	\item The semantic space includes meaning-based relationships among labels in the form of 
real-valued word vectors. 
\end{itemize}

\begin{figure}[t!]
	\centering
	\vspace{-0.3cm}
	\includegraphics[width=3.0in]{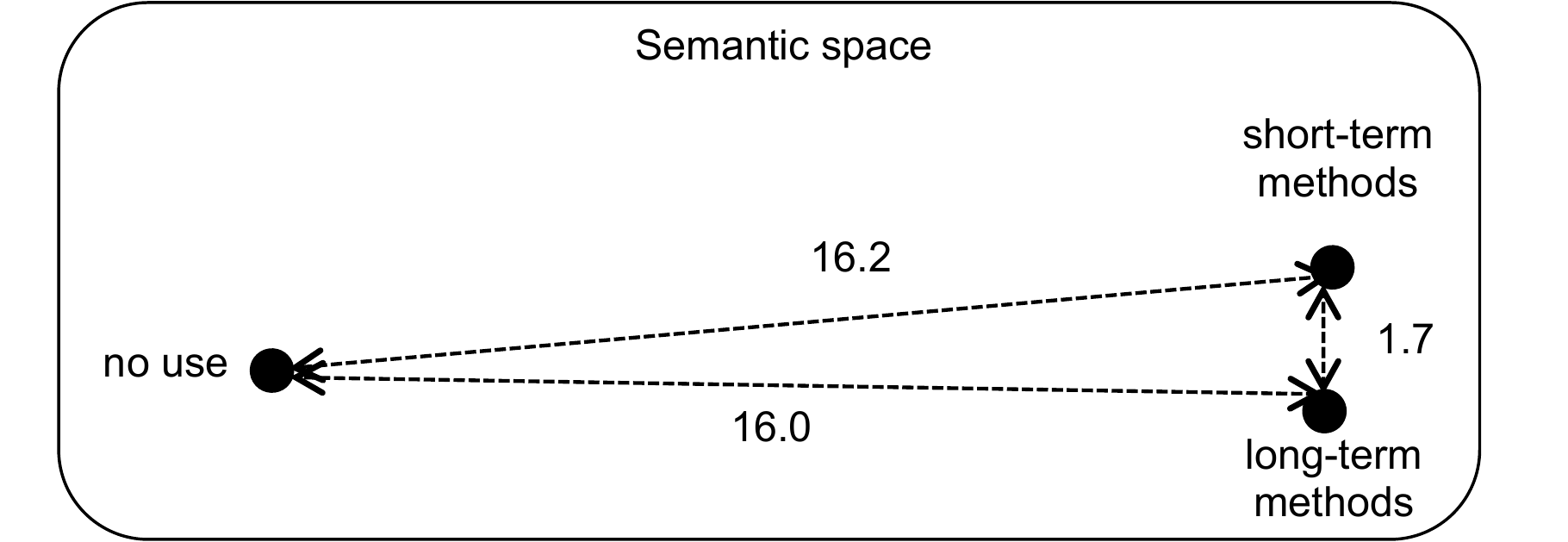}
	\vspace{-0.4cm}
	\DeclareGraphicsExtensions.
	\caption{Pairwise squared Euclidean distances in between the vector representations of labels 
		in the semantic space. A smaller distance indicates a larger class affinity.}
	\label{cmc_label_position_ss}
	\vspace{-0.4cm}
\end{figure}

As shown in Fig.~\ref{supervised_block}, traditional supervised learning operates in the feature 
space. It uses the features to model the data-label relationship. On the other hand, SECRET not only 
uses data available in the feature space, but also integrates meaning-based relationships among 
labels (semantic space) into the decision process, as shown in Fig.~\ref{secret_block}. SECRET requires vector representations of the training labels 
as an additional input, relative to the traditional supervised learning approach. Vector 
representations are obtained using semantic vector generation algorithms (see Section 
\ref{semantic_vector_models}) that are trained with a large number of documents. Depending on the 
available computational resources, SECRET can be implemented with either pre-trained semantic 
vectors that are available on the web \cite{glove_pretrained}, \cite{word2vec_pretrained}, 
\cite{word2vec_pretrained_biomedical}, or specially-trained semantic vectors obtained from a given 
corpus.  Neither implementation needs the involvement of an expert, unlike the case of labeling data 
in supervised learning.

The novelty of SECRET is that it enables an interaction between the two spaces while constructing the 
classifiers and regressors. In Fig.~\ref{secret_block}, the interaction is depicted by the arrow in between the two spaces. The hyperparameter values of the semantic (feature) space are not aimed
at maximizing the performance of the semantic space regressor 
(classifier), but that of the overall SECRET architecture. However, the interaction does not only 
take place during hyperparameter tuning. Unlike the traditional approaches, the classifier and 
regressor do not make individual decisions. Both provide confidence scores for each label. This 
information is used by SECRET to predict the label for a new query data instance. We explain each 
block next.

\subsection{Data Processing} \label{secret_process}
\begin{figure*}[t!]
	\centering
	\subfloat[]{\includegraphics[width=6.2in]{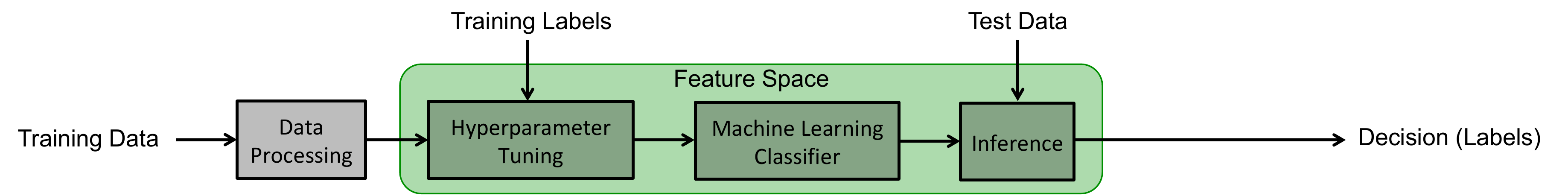}%
		\label{supervised_block}}
	\vspace{-0.4cm}
	\hfil
	\subfloat[]{\includegraphics[width=6.2in]{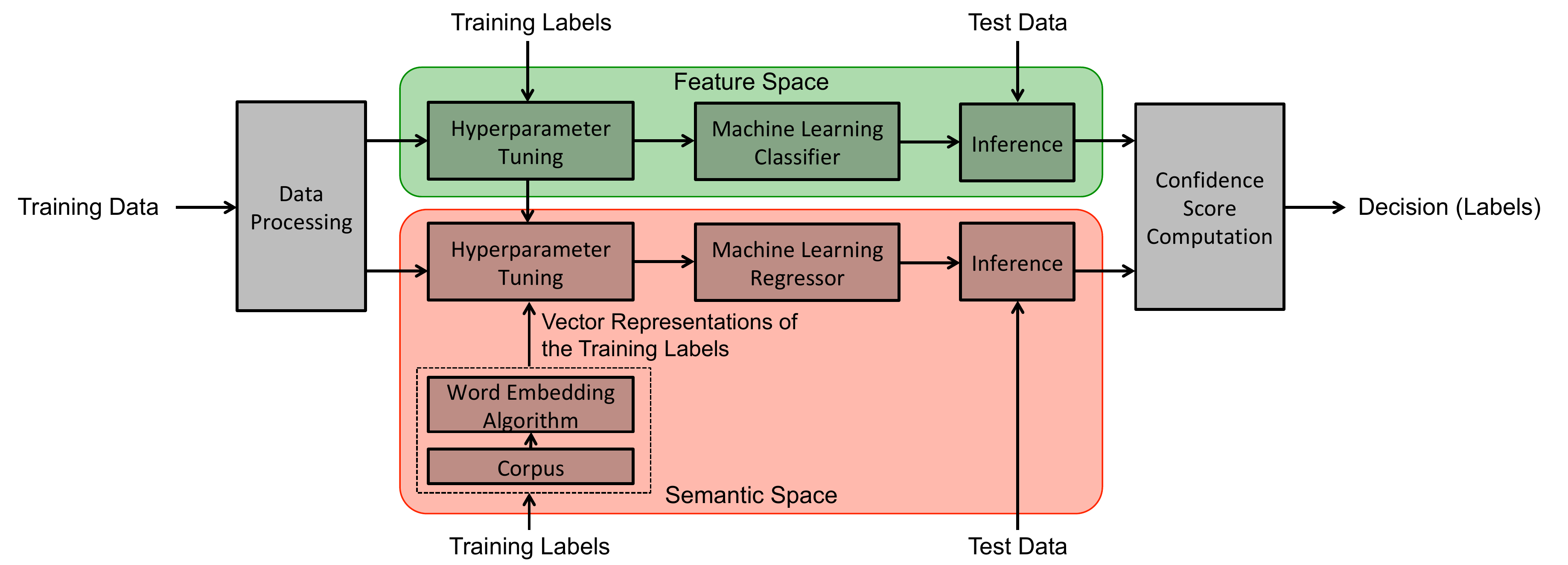}%
		\label{secret_block}}
	\vspace{-0.2cm}
	\hfil
	
	\caption{Architectures: (a) traditional supervised learning and (b) SECRET.}
	\vspace{-0.4cm}
	\label{block}
\end{figure*}

Data processing is an important part of any ML decision process. Data in the raw form require: 
(i) denoising \cite{motwani2004survey}, \cite{joshi2013survey}, \cite{kandaswamy2005removal}, 
\cite{mohan2014survey}, (ii) outlier elimination \cite{hodge2004survey}, (iii) feature extraction 
\cite{kelleher2015fundamentals}, (iv) feature encoding \cite{kelleher2015fundamentals}, and 
(v) normalization/standardization \cite{kelleher2015fundamentals}.
SECRET targets these operations in the `Data Processing' block in Fig.~\ref{secret_block}. For more 
details, readers are referred to the cited references.

\subsection{Hyperparameter Tuning} 
The selected set of hyperparameter values has a direct impact on classification/regression 
performance. In this work, we adopt Bayesian 
optimization for reasons mentioned in Section \ref{hyperparameter_tuning}.

By integrating exploration and 
exploitation, Bayesian optimization outputs the set of hyperparameter values that maximizes the 
optimization goal function. This function indicates the overall performance of the chosen supervised 
ML algorithm. Therefore, it guides Bayesian optimization to find the appropriate set of
hyperparameter values in order to enhance the performance of real-world decision processes. The pseudocode for the hyperparameter tuning stage is shown in Algorithm \ref{hyp_pseudocode}. 
Following preprocessing of training and validation data, the Gaussian Process (GP) of Bayesian 
optimization is initialized. Bayesian optimization takes hyperparameters (as variables, not their 
values), their ranges, and optimization goal function as input.  The hyperparameters depend on the 
chosen ML algorithm. For example, whereas the total number of trees may be a hyperparameter for the 
random forest (RF) algorithm, the number of layers and neurons in each layer may be hyperparameters 
for the multi-layer perceptron (MLP) algorithm. The optimization goal function reflects the purpose 
of the task being performed. Depending on whether SECRET is implemented on top of a traditional 
supervised (feature space) classifier or built from the ground up, the optimization 
goal function takes into account either available feature and semantic space hyperparameter 
values or both semantic and feature space hyperparameter values. The function outputs performance 
metrics, such as accuracy, F1 score, etc., based on training and validation data. We implement SECRET 
on top of a traditional supervised classifier in order to compare it with classification algorithms 
from the literature. Therefore, in Algorithm \ref{hyp_pseudocode}, the feature and semantic space 
algorithms are trained with already assigned hyperparameter values and acquisition function outputs 
(semantic space hyperparameter values), respectively. Then the feature and semantic space confidence 
scores are calculated and labels are assigned (see Section \ref{secret_decision} for details).  The optimization goal function output (classification performance on validation set) is used to update the beliefs and obtain 
the next set of semantic space hyperparameter values. The above process is repeated with this new 
set.  When the maximum number of allowed iterations (\textit{BOiter}) is reached, the process is 
stopped and the set of hyperparameter values (\textit{SSHyp}) that leads to 
the highest validation set performance is selected for application to the test set.

\subsection{Training of ML Models and Inference} \label{secret_training}

The data-label relationships exist in different forms in the feature and semantic spaces and
are captured through ML algorithms. The feature space does not take into account the meaning-based 
relationships among labels.  However, the semantic space takes into account the affinity and 
dissimilarity information between labels that is captured in a vector form. Hence, whereas 
the feature space decision process maps data to the label with the help of a classifier, the semantic 
space decision process relies on a regressor. The choice of the regressor has a direct impact on 
SECRET's performance. Thus, for a fixed feature space classifier, we carry out performance 
analyses with various regressors on the training and validation data and select the one that
best maps data to labels. After finding the best set of hyperparameter values in both spaces 
through joint optimization, we train the ML algorithms. We train the feature space 
classifier with the selected hyperparameter values, training data, and training labels. 
Then we train the  semantic space regressor with the selected hyperparameter values, training data, and
vector representations of the labels. Following the training stage, we perform inference on
the test data and obtain the confidence scores. We show the operations corresponding to this stage 
in lines 3 through 6 in Algorithm \ref{dec_pseudocode}.

\subsection{Confidence Score Computation and Decision} \label{secret_decision}
\begin{figure}[t!]
	\centering
	\includegraphics[width=3.0in]{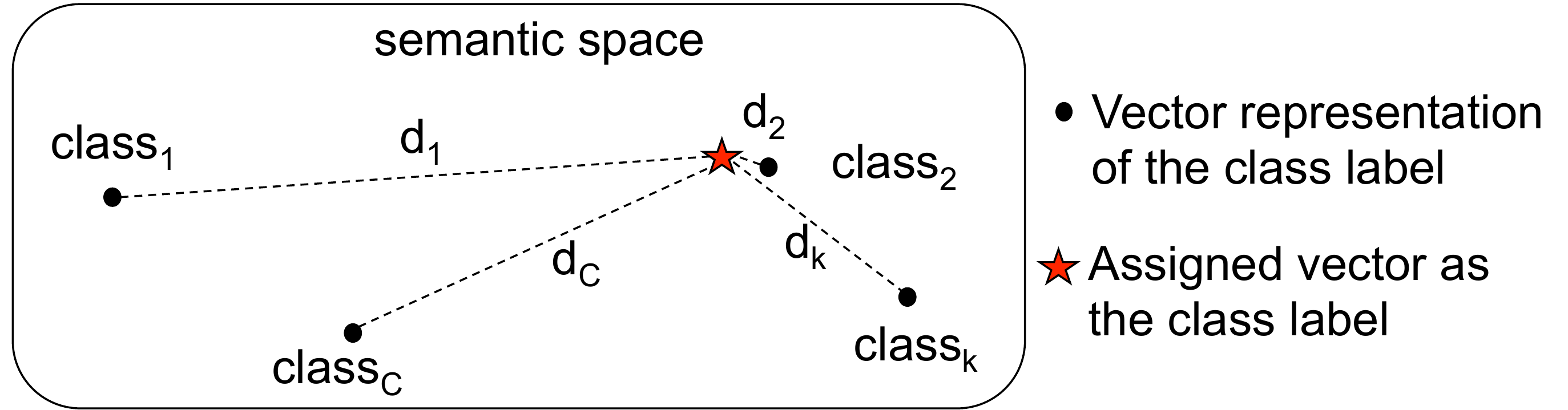}
	\vspace{-0.4cm}
	\DeclareGraphicsExtensions.
	\caption{Confidence score computation in the semantic space. Confidence score is
calculated for each $k$, where $k \in [1,C]$ and $C$ represents the total number of classes.}
	\label{ss_conf}
	\vspace{-0.4cm}
\end{figure}

The inference stage outputs the confidence scores for each data instance for both spaces. 
The feature space confidence score ($FSConf$) computation depends on the chosen ML algorithm.
For example, the confidence score of the RF classifier is computed as the average class probabilities 
of all trees.  The class probability for each tree is computed as the fraction of the samples that 
belong to the class of interest in a leaf.  However, the confidence score of the MLP classifer is the 
same as the output of the activation function in the outermost (final) layer. On the other hand, as 
shown in Fig.~\ref{ss_conf}, the semantic space confidence score, $SSConf$, is based on Euclidean 
distance, in line with the main motivation behind semantic vector representations. The distance between the vector representation of the class label and the assigned vector is inversely proportional to $SSConf$. As shown in 
Eq.~\ref{ss_conf_eq}, $SSConf$ is computed through the normalized inverse ratio of the squared 
distance between the assigned vector and label vector. In Algorithm \ref{dec_pseudocode}, the 
corresponding operations are shown in lines 7 through 9. $D$, $V$, and $C$ represent the dimension 
of the semantic word vector, semantic vector of the class label, and total number of classes, 
respectively. $\epsilon$ refers to additive shift. In line $9$, $\epsilon$ is used to avoid 
divergence of the algorithm when the assigned vector (regressor output) overlaps with the vector of 
the class label. The value of $\epsilon$ needs to be smaller than the minimum difference between 
vectors of the assigned label and the class label.  However, in the hyperparameter tuning stage 
(line $14$ in Algorithm \ref{hyp_pseudocode}), $\epsilon$ is assigned $0$ and divergence is permitted 
to avoid overfitting.  Since divergence blocks label assignment, an overlap between the assigned 
and class labels degrades classification performance on the validation set significantly. Therefore, 
the hyperparameters corresponding to this case are not selected (line $17$ through line $19$ in 
Algorithm \ref{hyp_pseudocode}) and overfitting is avoided. In summary, while assigning $0$ to 
$\epsilon$ in the hyperparameter tuning stage is beneficial for preventing overfitting, a nonzero 
value in the decision making process is needed to avoid divergence.

\begin{algorithm} [t!]
	\caption{\textbf{} SECRET - Hyperparameter Tuning}\label{hyp_pseudocode} 
	\textbf{Input:} \\
	$\textit{DataTr}$: Training data\\
	$\textit{DataVal}$: Validation data\\
	$\textit{LabelTr}$: Labels corresponding to training data\\
	$\textit{LabelVal}$: Labels corresponding to validation data\\
	$\textit{FSHyp}$: Feature space hyperparameter values\\
	$\textit{V}$: Vector representations of the labels\\
	
	\textbf{Output:} \\
	$\textit{SSHyp}$: Semantic space hyperparameter values\\
	
	\begin{algorithmic}[1]
		\State $Preprocess$ $DataTr$, See Section \ref{secret_process}
		\State $Preprocess$ $DataVal$, See Section \ref{secret_process}
		\State $ACC_{max} \gets \textit{0}$ 
		\State $SSHyp \gets \textit{null}$ 
		\State $Initialize$ $GP$ $of$ $Bayesian$ $Optimization$
		\State \textbf{for} $i=1,..., BOiter$
		\State $ $ $ $ $ $ $ $ $ $ $FSModel \gets \textit{FSClassifier(DataTr,LabelTr,FSHyp)}$
		\State $ $ $ $ $ $ $ $ $ $ $FSConf \gets \textit{FSModel(DataVal)}$ 
		\State $ $ $ $ $ $ $ $ $ $ $SSHyp_i \gets \textit{Exploration+Exploitation}$
		\State $ $ $ $ $ $ $ $ $ $ $SSModel_i \gets \textit{SSRegressor(DataTr,LabelTr,SSHyp$_i$)}$
		\State $ $ $ $ $ $ $ $ $ $ $SSOut \gets \textit{SSModel$_i$(DataVal)}$
		\State $ $ $ $ $ $ $ $ $ $ \textbf{for} $j=1,..., \# instances$
		\State $ $ $ $ $ $ $ $ $ $ $ $ $ $ $ $ $ $ $ $ \textbf{for} $k=1,..., C$
		\State $ $ $ $ $ $ $ $ $ $ $ $ $ $ $ $ $ $ $ $ $ $ $ $ $ $ $ $ $ $ $SSConf_{jk} =\frac{\frac{1}{\sum_{l=1}^{D}(V_{k}(l)-SSOut_{jk}(l))^2}}{\sum_{m=1}^{C}\frac{1}{\sum_{j=1}^{D}(V_{m}(l)-SSOut_{jk}(l))^2}}$ 
		
		\State $ $ $ $ $ $ $ $ $ $ $LabelValPred \gets \textit{argmax$_{class}$(avg(FSConf,SSConf))}$
		\State $ $ $ $ $ $ $ $ $ $ $ACC_i \gets \textit{ACC(LabelVal,LabelValPred)}$
		
		\State $ $ $ $ $ $ $ $ $ $ \textbf{if} $ACC_i>ACC_{max}$
		\State $ $ $ $ $ $ $ $ $ $ $ $ $ $ $ $ $ $ $ $ $ACC_{max}=ACC_i$
		\State $ $ $ $ $ $ $ $ $ $ $ $ $ $ $ $ $ $ $ $ $SSHyp=SSHyp_i$
		
		\State $ $ $ $ $ $ $ $ $ $ $Update$ $GP$
		\State \textbf{return} $SSHyp$ 
		
	\end{algorithmic}
	
\end{algorithm}

\begin{equation}
\begin{aligned}
SSConf_k = \frac{\frac{1}{d_k^2 + \epsilon }}{\sum_{i=1}^{C}{\frac{1}{d_i^2 + \epsilon}}},
k \in [1,C],\\
\text{where C\text{: Total number of classes}} , \\
d \text{: Distance between assigned vector and class label},\\
\epsilon \text{: Additive shift}.
\end{aligned}
\label{ss_conf_eq}
\end{equation}

\begin{figure}[t!]
	\centering
	\includegraphics[width=2.5in]{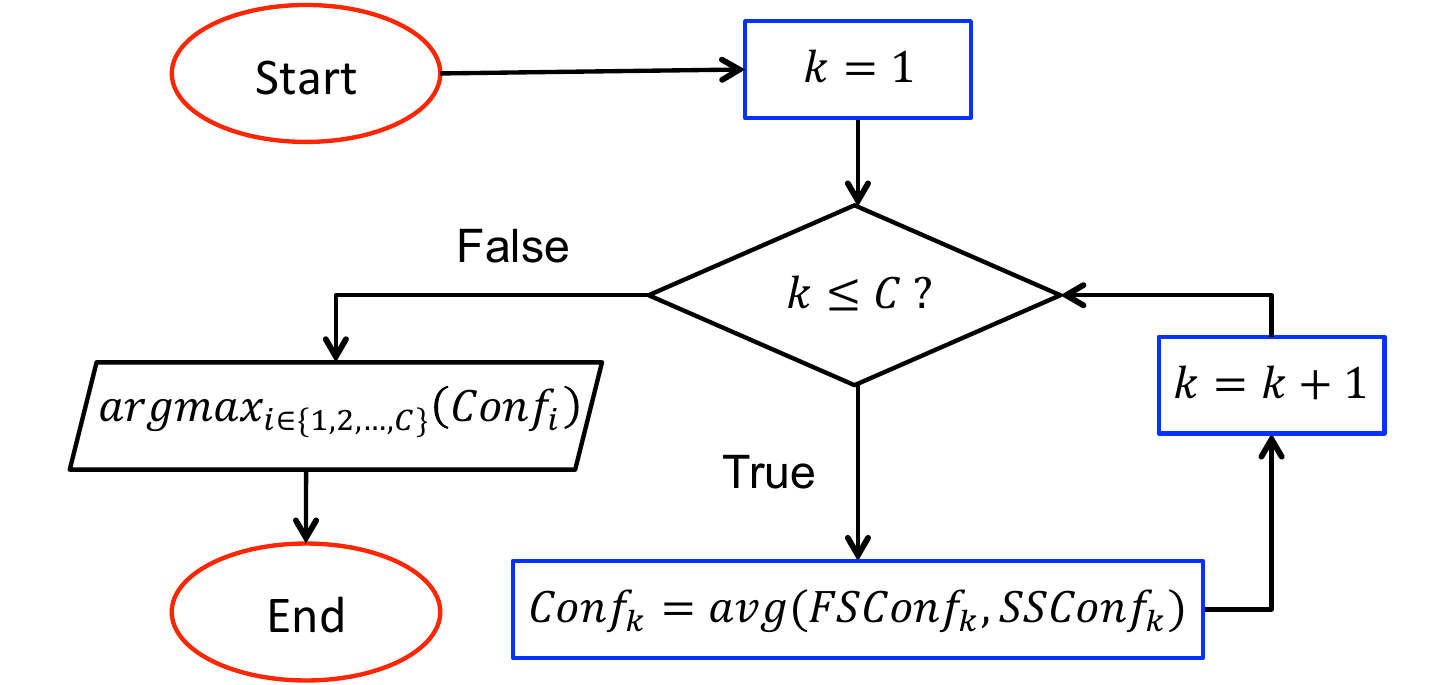}
	\DeclareGraphicsExtensions.
	\vspace{-0.4cm}
	\caption{Overall confidence score computation and decision-making stages of SECRET.}
	\label{conf}
	\vspace{-0.4cm}
\end{figure}

The overall confidence score is computed by taking the average of $FSConf$ and $SSConf$ for each 
class (Fig. \ref{conf}). In the decision stage, each data instance in the test set is assigned the label of the class 
that has the highest overall confidence score (line $10$ in Algorithm \ref{dec_pseudocode}). 
Following the labeling stage, SECRET's classification 
performance is assessed through accuracy ($ACC$) and F1 score metrics computed using 
Eq.~\ref{acc} and Eq.~\ref{f1}, respectively. Accuracy depicts the ratio of the number of correctly 
classified instances and the total number of instances. However, the F1 score indicates the 
fraction of correctly classified instances for each class within the dataset. 

\begin{equation} 
\begin{aligned}
ACC = \frac{TP + TN}{TP+TN+FP+FN}\, \\
\text{where } TP \text{: True positive}, 
TN \text{: True negative}, \\
FP \text{: False positive},
FN \text{: False negative}. \\
\end{aligned}
\label{acc}
\end{equation}

\begin{equation}
\begin{aligned}
F1 = \frac{\sum_{i=1}^{C}{2*PREC_i*REC_i/(PREC_i+REC_i)}}{C},\\
\text{where } PREC = \frac{TP}{TP+FP}, 
REC = \frac{TP}{TP+FN},\\ 
C \text{: Total number of classes},\\
PREC \text{: Precision}, 
REC \text{: Recall}. \\
\end{aligned}
\label{f1}
\end{equation}

\section{Experimental Results and Discussion} \label{results}
In this section, we present the experimental results for SECRET and provide comparisons with
traditional supervised classifiers and ensemble methods. Then, we analyze the effect of 
feature-semantic space variations on SECRET's classification performance.

\subsection{Datasets} \label{datasets}  
SECRET's flexible design ensures applicability to a broad spectrum of real-world classification 
tasks. We analyze its performance on datasets for ten different applications, ranging 
from biomedical disease diagnosis to sonar-based object detection. Table \ref{dataset} describes 
these datasets and their characteristics. The datasets are taken from the UCI Machine Learning 
Repository \cite{Dua:2017}. They focus on the classification task. 

\begin{algorithm} [t!]
	\caption{\textbf{} SECRET - ML Training - Inference - Decision}\label{dec_pseudocode} 
	\textbf{Input:} \\
	$\textit{DataTrVal}$: Training and validation data\\
	$\textit{DataTe}$: Test data\\
	$\textit{LabelTrVal}$: Labels corresponding to training and validation data\\
	$\textit{LabelTe}$: Labels corresponding to test data\\
	$\textit{FSHyp}$: Feature space hyperparameter values\\
	$\textit{SSHyp}$: Semantic space hyperparameter values\\
	$\textit{V}$: Vector representations of the labels\\
	
	\textbf{Output:} \\
	$\textit{ACC}$: SECRET's accuracy on the test set\\
	$\textit{F1}$: SECRET's F1 score on the test set\\
	
	\begin{algorithmic}[1]
		
		\State $Preprocess$ $DataTrVal$, See Section \ref{secret_process}
		\State $Preprocess$ $DataTe$, See Section \ref{secret_process}
		\State $FSModel \gets \textit{FSClassifier(DataTrVal,LabelTrVal,FSHyp)}$
		\State $FSConf \gets \textit{FSModel(DataTe)}$ 
		\State $SSModel \gets \textit{SSRegressor(DataTrVal,LabelTrVal,SSHyp)}$
		\State $SSOut \gets \textit{SSModel(DataTe)}$
		\State \textbf{for} $j=1,..., \# instances$
		\State $ $ $ $ $ $ $ $ $ $ \textbf{for} $k=1,..., C$
		\State  $ $ $ $ $ $ $ $ $ $ $ $ $ $ $ $ $ $ $ $ $SSConf_{jk} =\frac{\frac{1}{\sum_{l=1}^{D}(V_{k}(l)-SSOut_{jk}(l))^2+\epsilon}}{\sum_{m=1}^{C}\frac{1}{\sum_{j=1}^{D}(V_{m}(l)-SSOut_{jk}(l))^2+\epsilon}}$ 
		
		\State $LabelTePred \gets \textit{argmax$_{class}$(avg(FSConf,SSConf))}$
		\State $ACC \gets \textit{ACC(LabelTe,LabelTePred)}$
		\State $F1 \gets \textit{F1(LabelTe,LabelTePred)}$
		
		\State \textbf{return} $ACC,F1$ 
		
	\end{algorithmic}
\end{algorithm}

The UCI Connectionist Bench Dataset is based on sonar signals that are reflected from a rock or metal cylinder. It discriminates between these two obstacles. The UCI Indian Liver Patient Dataset is composed of patient records, such as age, gender, total bilirubin, total protein, albumin, etc. It is aimed at diagnosing liver disease. The UCI Breast Cancer Wisconsin Dataset is built using images of a fine needle aspiration of the breast. It focuses on classifying the cell nucleus as malignant or benign. The Statlog Dataset is composed of physiologic signal and demographic information of patients. It is aimed at predicting the absence or presence of heart disease. The UCI Contraceptive Method Choice Dataset is composed of demographic and socio-economic information of married women. It is aimed at identifying their contraceptive method choices. The UCI Lymphography Dataset includes features extracted from lymphography images. It is aimed at classifying different 
types of lymph nodes. The UCI Nursery Dataset includes parental occupation, child's nursery condition, family structure, and the family's social, health, and financial status as features. It is targeted at ranking of nursery school applications. The UCI Cardiotocography Dataset is formed using cardiotocography features that are based on fetal 
heart rate and uterine contraction. It focuses on classification of ten different fetal morphologic patterns. The UCI Chess Dataset is built using the king-rook and rook positions on a chessboard. It is targeted at the depth of a win. The UCI Letter Recognition Dataset is formed using black-and-white image pixels of letters of the English alphabet. It is targeted at classifying each of the 26 letters. 

\begin{table*}[t!]
	\caption{Datasets and Their Characteristics}
	\centering
	\begin{tabular}{p{2in} c c c c p{2in}}
		
		\hline
		Dataset& Abbreviation& \texttt{\#} Instances& \texttt{\#} Features& \texttt{\#} Classes& Class Labels\\[0.5ex]
		
		\hline\hline
		UCI Connectionist Bench (Sonar, Mines vs. Rocks) Dataset&sonar&208&60&2&Rock, Metal cylinder\\[0.5ex]
		UCI ILPD (Indian Liver Patient Dataset)&liver&583&10&2&Liver patient, Not liver patient\\[0.5ex]
		UCI Breast Cancer Wisconsin (Diagnostic) Dataset&wdbc&569&30&2&Benign, Malignant\\[0.5ex]
		UCI Statlog (Heart) Dataset&heart&270&13&2&Absence, Presence\\[0.5ex]
		UCI Contraceptive Method Choice Dataset&cmc&1473&9&3&No use, Short-term methods, Long-term methods\\[0.5ex]
		UCI Lymphography Dataset&lymph&148&18&4&Normal find, Metastases, Malign lymph, Fibrosis\\[0.5ex]
		UCI Nursery Dataset&nursery&12960&8&5&Not recommended, Recommended, Very recommended, Priority, Special Priority\\[0.5ex]
		UCI Cardiotocography Dataset&cardio&2126&21&10&Calm sleep, REM sleep, Calm vigilance, Active vigilance, Shift pattern, Stress situation, Vagal stimulation, Largely vagal stimulation, Pathological state, Suspect pattern\\[0.5ex]
		UCI Chess (King-Rook vs. King) Dataset&chess&28056&6&18&Draw, Zero, One, Two, Three, Four, Five, Six, Seven, Eight, Nine, Ten, Eleven, Twelve, Thirteen, Fourteen, Fifteen, Sixteen\\[0.5ex]
		UCI Letter Recognition Dataset&letter&20000&16&26&A, B, C, D, E, F, G, H, I, J, K, L, M, N, O, P, Q, R, S, T, U, V, W, X, Y, Z\\[0.5ex]	
		\hline
		\hline	
	\end{tabular}
	\label{dataset}
\end{table*}

\begin{figure*}[t!]
	\centering
	\includegraphics[width=6.5in]{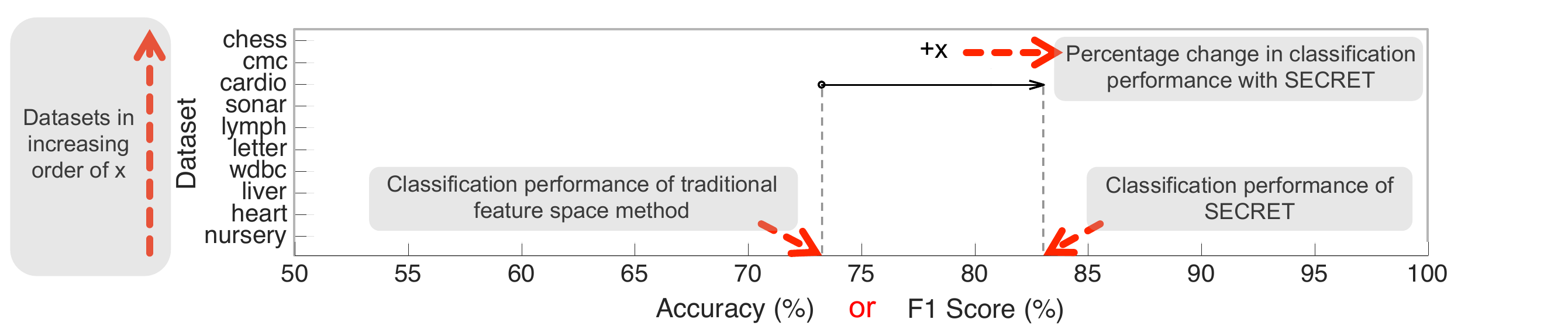}
	\vspace{-0.4cm}
	\DeclareGraphicsExtensions.
	\caption{Legend for experiments that compare traditional approches with SECRET.}
	\vspace{-0.4cm}
	\label{secret_legend}
	
\end{figure*}

\subsection{Supervised Classifier vs. SECRET} \label{secret_vs_supervised} 
We hypothesize that feature space information is not the only source of information
that can be used for classification. In order to test this hypothesis,
we compare the classification performance of the supervised classifier with
that of SECRET. The supervised classifier uses feature space information to model
the data-label relationship and predict labels of unlabeled data instances. On the 
other hand, SECRET fuses the feature and semantic space information to predict labels. 
By analyzing the two approaches, we aim to identify the impact of semantic space information
on classification performance.  In order to minimize dependency on an ML algorithm, we use
two classifiers (RF and MLP) and their regressor versions. Decision trees in RF are 
information-based; however, MLP is error-based. Second, in order to avoid a biased evaluation 
of classification performance, we compute both accuracy and F1 scores by comparing the 
predicted labels with actual ones in the test set. Accuracy reflects the percentage of correctly 
classified samples within the test set. However, the F1 score incorporates precision and recall 
values, which are computed from the false positive, false negative, and true positive values for each class, and 
then taking the average.  Therefore, accuracy and F1 score assess the classification performance 
from different perspectives. 

We implement SECRET with a Bayesian optimization framework \cite{bo_framework} (used for determining the number of neurons in an MLP with a single hidden layer and the number of trees in RF), 
scikit-learn \cite{scikit-learn}, and 50-dimensional GloVe vectors \cite{glove_pretrained} 
(pretrained with Wikipedia 2014 + Gigaword 5). In the case of a label with more than one word, we take the average of the word vectors corresponding to the words within the label to find the overall label vector. We compute the semantic space confidence score 
using the equations shown in line $14$ of Algorithm \ref{hyp_pseudocode} and line $9$ of Algorithm 
\ref{dec_pseudocode}.  Based on our analyses (see the conditions mentioned in 
Section~\ref{secret_decision}), $\epsilon$ is assigned $10^{-200}$ in the experiments. For 
classification performance analyses, we use stratified $10$-fold sampling for each dataset and 
report the average accuracy and F1 score values.

Fig.~\ref{secret_legend} shows the legend for the plots that depict classification performance of 
the traditional feature space approach and SECRET. The starting points of the arrows indicate 
accuracy or F1 scores of the feature-space classifiers. The ending points indicate the impact 
of including semantic space information on classification. The numbers 
above/below the arrows show the percentage improvement. The arrow sizes are scaled accordingly. 
Dataset names are ordered based on the amount of change in classification performance 
caused by SECRET. The dataset on top shows the largest change. The amount of change decreases from top to bottom.

\begin{figure*}[t!]
	\centering
	\subfloat[]{\includegraphics[width=6.0in]{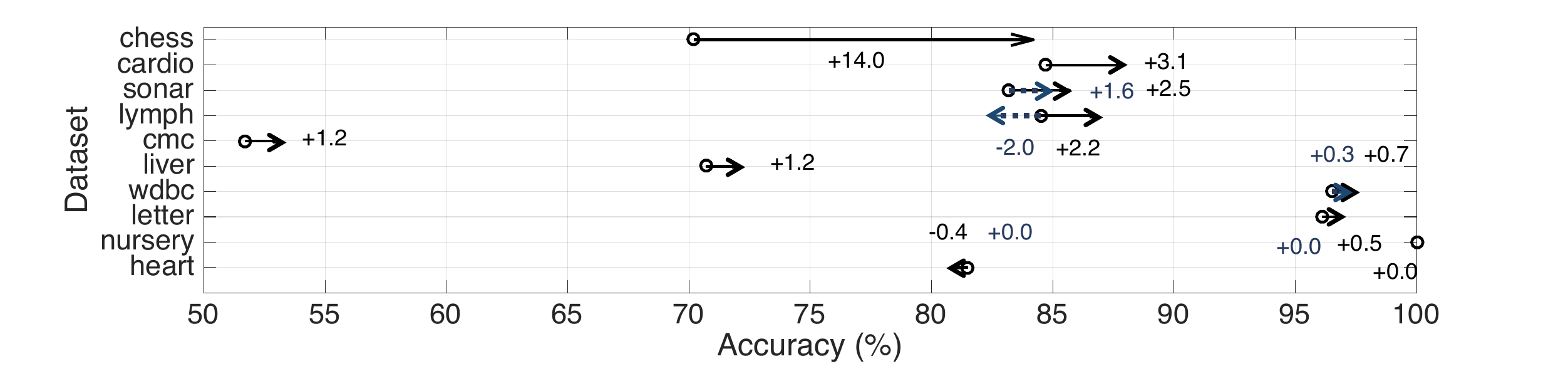}%
		\label{fs_mlp_secret_acc}}
	\vspace{-0.4cm}
	\hfil
	\subfloat[]{\includegraphics[width=6.0in]{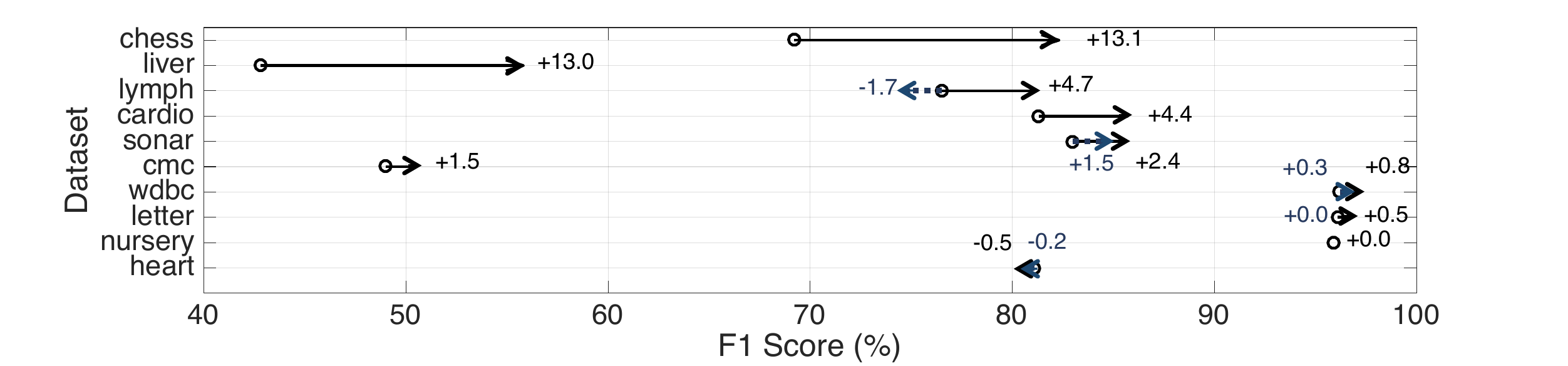}%
		\label{fs_mlp_secret_f1}}
	\vspace{-0.2cm}
	\hfil
	\caption{SECRET's (a) accuracy and (b) F1 score improvements over the traditional MLP 
		(feature-space) classifier. SECRET uses MLP as the feature space classifier and 
		RF/MLP as the semantic space regressor. Black arrows indicate when the RF regressor 
                is used, whereas the dark blue and dashed arrows correspond to the MLP regressor.}
	\vspace{-0.4cm}
	\label{fs_mlp_secret}
\end{figure*}

\begin{figure*}[t!]
	\centering
	\subfloat[]{\includegraphics[width=6.0in]{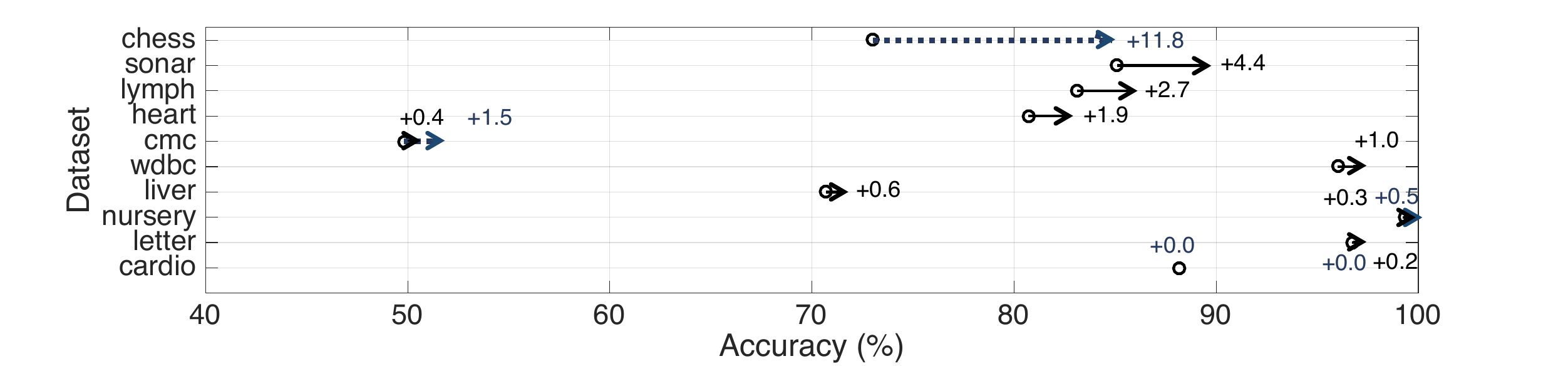}%
		\label{fs_rfc_secret_acc}}
	\vspace{-0.4cm}
	\hfil
	\subfloat[]{\includegraphics[width=6.0in]{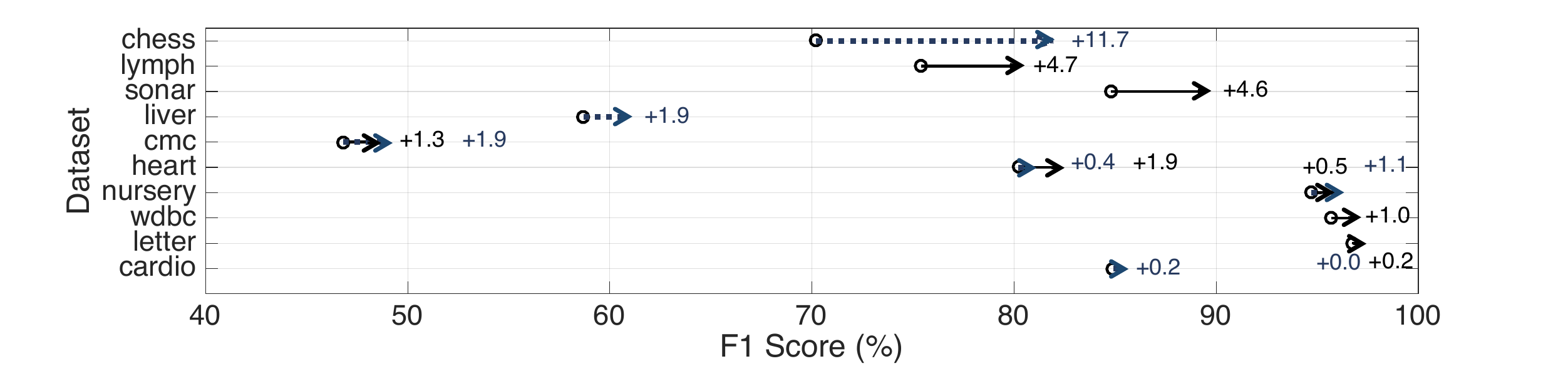}%
		\label{fs_rfc_secret_f1}}
	\vspace{-0.2cm}
	\hfil
	\caption{SECRET's (a) accuracy and (b) F1 score improvements over the traditional RF 
		(feature-space) classifier. SECRET uses RF as the feature space classifier and 
                MLP/RF as the semantic space regressor. Black arrows indicate when the MLP regressor 
                is used, whereas the dark blue and dashed arrows correspond to the RF regressor.}
	\vspace{-0.4cm}
	\label{fs_rfc_secret}
\end{figure*}

Fig.~\ref{fs_mlp_secret} shows the accuracy and F1 scores of the traditional MLP classifier and
SECRET with the format shown in Fig.~\ref{secret_legend}. In this case, SECRET integrates semantic 
information into the MLP classifier with the help of either an RF or MLP regressor. It chooses the 
type of regressor based on validation set performance, builds the overall classifier using both the 
semantic and feature spaces, and makes the final decision on the test labels. The color of arrows in 
Fig.~\ref{fs_mlp_secret}
indicates the chosen regressor type. If both black and dark blue colored arrows are shown for a
dataset, then the validation set classification performance is inconclusive (both regressors perform equally well on the validation set with less than or equal to 1.0\% difference in accuracy 
and F1 score values) in determining the 
better regressor. We present results with both regressors. 
The chess dataset has 14.0\% and 13.1\% improvements in accuracy and F1 scores, respectively. 
While the liver dataset has 
1.2\% improvement in accuracy, it has the second highest F1 score improvement of 13.0\% among all 
datasets. This shows the importance of analyzing classification performance from different 
perspectives. When there is class imbalance, F1-macro score shines light on classification
performance of the minority class. As in the case of the liver dataset, biomedical disease
diagnosis datasets are likely to be imbalanced. Although the main goal is to detect the
“disease”, in general, the `healthy' class has more datapoints than the “disease” one. Class
imbalance might affect performance. However, it does not only affect the performance of the
feature space classifier, but also that of SECRET. We upsampled the minority class with the SMOTE 
method and repeated the experiments to test this point. While the feature space 
resulted in 60.3\% accuracy and 58.1\% F1 score, SECRET led to 69.8\% accuracy and 65.0\% F1 score. 
Although upsampling improved classification performance of both approaches, SECRET's accuracy and 
F1 score dominated the traditional feature space classifier by 9.5\% and 6.9\%, respectively. 
Moreover, except for the lymph and heart datasets, we observe that the arrows point to the right, 
indicating that SECRET improves accuracy as well as the F1 score. More specifically, use of the RF or MLP regressor, choice depending on the validation set performance, in the semantic space leads to classification performance improvement in nine out of 
ten and eight out of ten datasets, respectively. The amount of improvement depends on dataset 
characteristics, feature space classifier, and chosen semantic space regressor. 

Fig.~\ref{fs_rfc_secret} shows the classification performance of the traditional RF classifier 
and SECRET implemented with an MLP or RF regressor. In this experimental setup, SECRET shows 
11.8\% accuracy and 11.7\% F1 score improvements for the chess dataset. Moreover, SECRET achieves 
4.4\% accuracy and 4.6\% F1 score improvements for the sonar dataset. For the lymph dataset, we 
observe a 2.7\% accuracy and 4.7\% F1 score improvement with the MLP regressor. For the liver 
dataset, as in the case of Fig.~\ref{fs_mlp_secret}, a 1.9\% increase in the F1 score points to the 
positive impact of the semantic space information on decreasing the false 
positive and false negative rates, thus increasing the precision and recall values. None of the 
arrows in Fig.~\ref{fs_rfc_secret} point to the left, confirming the stable performance enhancement 
of SECRET that is independent of application type or dataset. 

Overall, the  results shown in Fig.~\ref{fs_mlp_secret} and Fig.~\ref{fs_rfc_secret} demonstrate 
classification performance improvement with SECRET. 

\begin{figure*}[t!]
	\centering
	\includegraphics[width=6.4in]{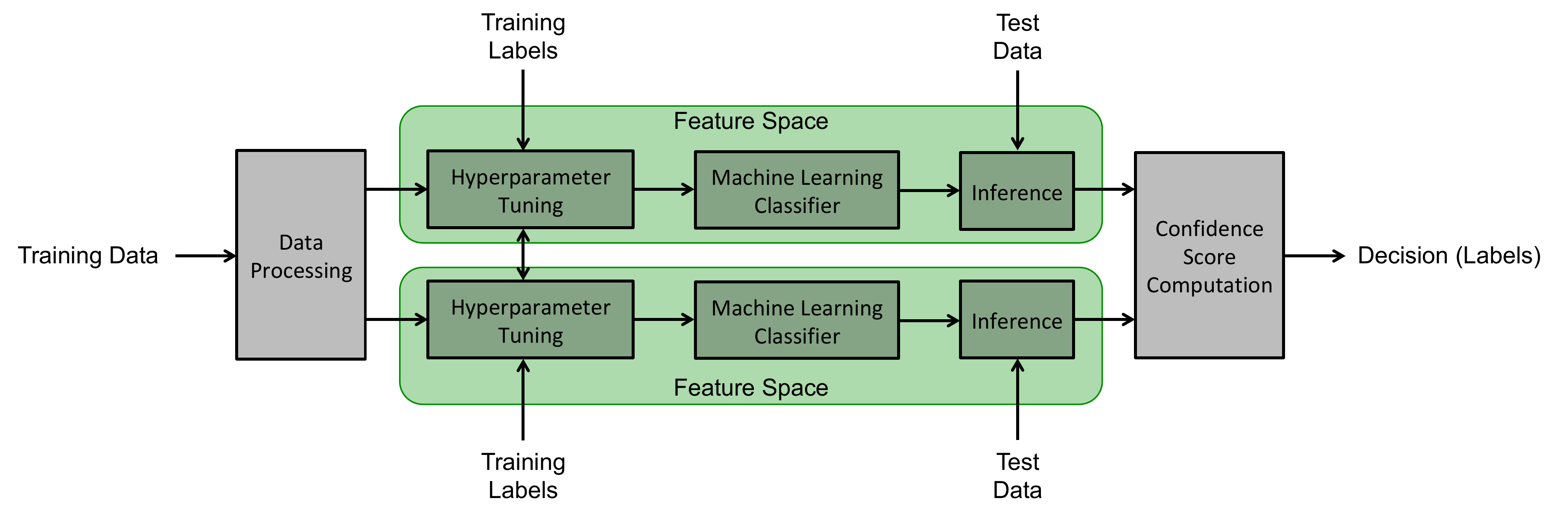}
	\DeclareGraphicsExtensions.
	\vspace{-0.4cm}
	\caption{Architecture of the traditional ensemble method.}
	\vspace{-0.4cm}
	\label{ensemble_block}
\end{figure*}

\begin{figure*}[t]
	\centering
	\subfloat[]{\includegraphics[width=6.0in]{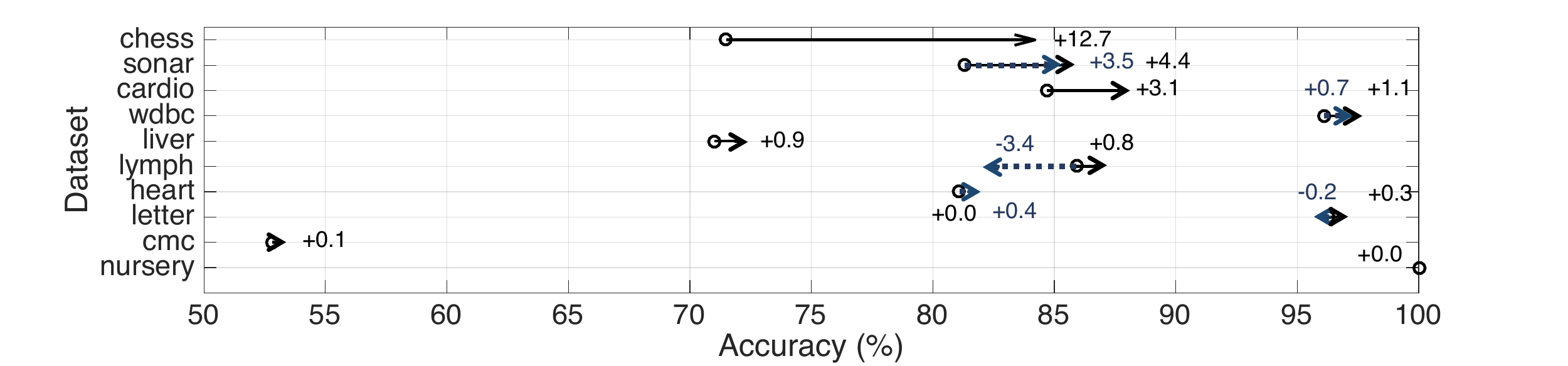}%
		\label{fs_fs_mlp_mlp_secret_acc}}
	\vspace{-0.4cm}
	\hfil
	\subfloat[]{\includegraphics[width=6.0in]{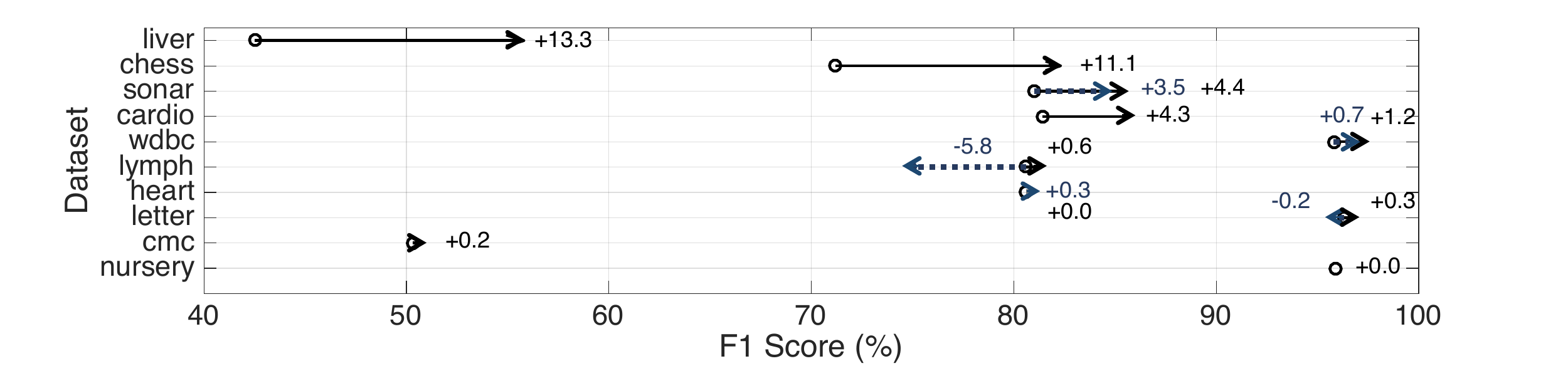}%
		\label{fs_fs_mlp_mlp_secret_f1}}
	\vspace{-0.2cm}
	\hfil
	\caption{SECRET's (a) accuracy and (b) F1 score improvements over the traditional MLP-MLP 
		ensemble in the feature space. SECRET uses MLP as the feature space classifier and RF/MLP as the 
		semantic space regressor. Black arrows indicate when the RF regressor is used, whereas the dark blue and 
		dashed arrows correspond to the MLP regressor.}
	\label{fs_fs_mlp_mlp_secret}
	\vspace{-0.4cm}
\end{figure*}

\subsection{Ensemble Method vs. SECRET} \label{secret_vs_ensemble} 
We saw in the previous section that SECRET outperforms traditional supervised ML classifiers. 
However, the traditional classifier also can be made more robust by using an ensemble method.
In this section, we compare ensemble methods with SECRET to show that the semantic space offers
a different type of information source that pays rich dividends.
In order to have a fair comparison between traditional ensemble methods and SECRET, we replace 
the red 'Semantic Space` block in Fig.~\ref{secret_block} with a 'Feature Space` block. The 
corresponding block diagram for the ensemble method is shown in Fig.~\ref{ensemble_block}. The
ensemble method is composed of only feature space classifiers. In the experiments, we provide 
the same amount of processing, hyperparameter tuning, and decision-making resources to the two 
approaches. The only difference is that only the feature space information is used in the ensemble 
method, whereas both the feature and semantic space information is used in SECRET. We analyze the 
ensembles (formed with MLP and RF algorithms) and compare them with SECRET next.

Fig.~\ref{fs_fs_mlp_mlp_secret} shows the accuracy and F1 scores of the traditional ensemble method 
and SECRET on the ten datasets. The ensemble is built using an MLP classifier whose performance is 
maximized with the best set of hyperparameter values. Then this classification performance is enhanced 
by combining the classifier with another MLP with hyperparameter values that maximize the overall 
performance of the ensemble. SECRET is built in the same way. However, the feature space classifier 
is replaced with a regressor that models the data and semantic vector relationship. 
Since the lymph dataset has the smallest size, the use of one regressor or the other leads to a 
significant classification performance degradation or some improvement. Due to this instability, we 
do not use the lymph dataset to come to a conclusion.  For seven datasets (chess, sonar, cardio, 
wdbc, liver, heart, and cmc), SECRET achieves a 0.1 to 12.7\% higher accuracy and a 
0.2 to 13.3\% higher F1 score relative to the ensemble method.  For the nursery dataset, both 
approaches show comparable classification performance. Also, as in the experiments described in 
Section \ref{secret_vs_supervised}, while the liver dataset has a 0.9\% increase in accuracy with 
SECRET, it obtains the highest F1 score improvement of 13.3\%.  Overall, as indicated by 
rightward-pointing arrows, SECRET can be seen to outperform the ensemble method. 

\begin{figure*}[t!]
	\centering
	\subfloat[]{\includegraphics[width=6.0in]{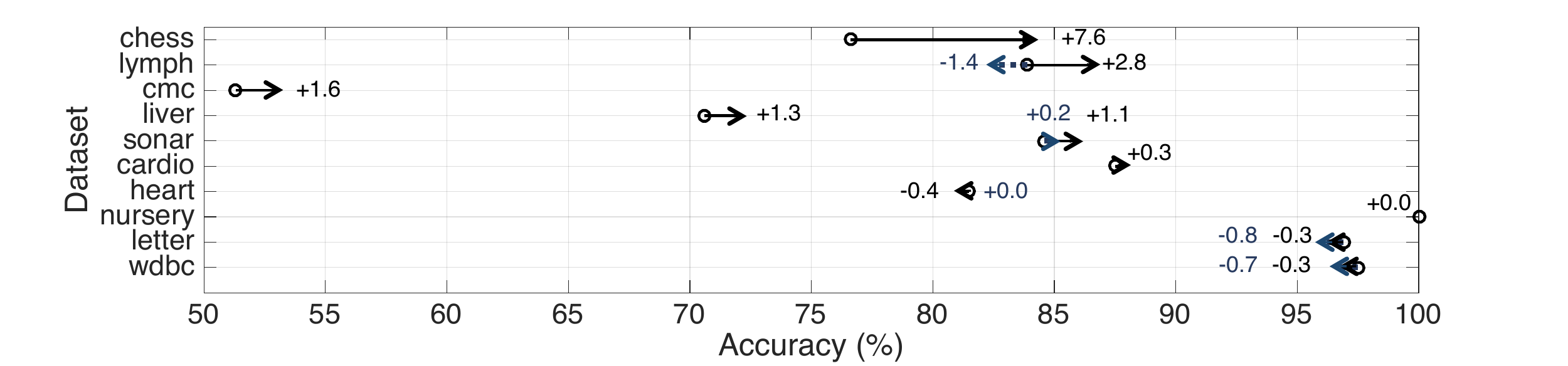}%
		\label{fs_fs_mlp_rfc_secret_acc}}
	\vspace{-0.4cm}
	\hfil
	\subfloat[]{\includegraphics[width=6.0in]{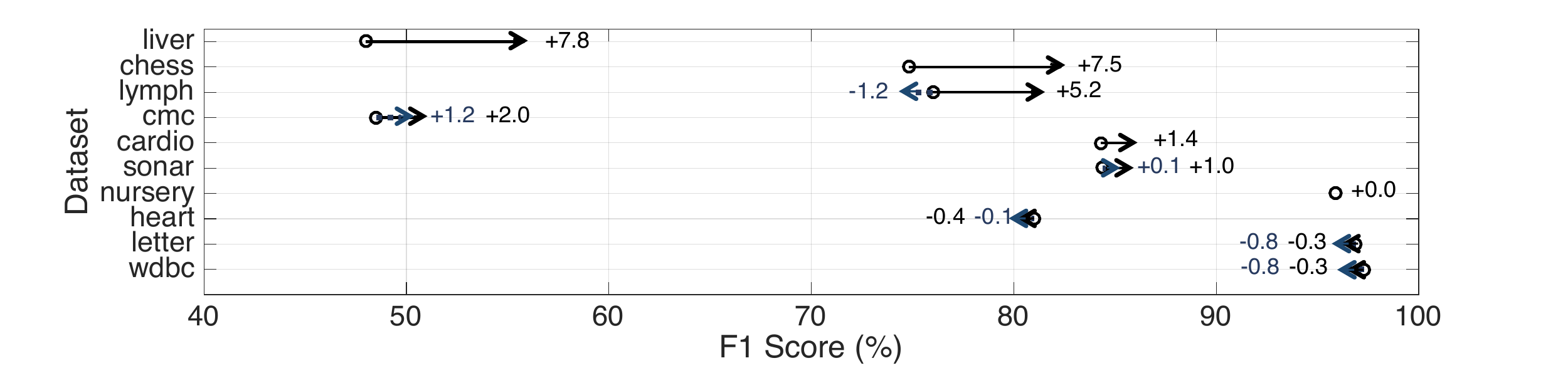}%
		\label{fs_fs_mlp_rfc_secret_f1}}
	\vspace{-0.2cm}
	\hfil
	\caption{SECRET's (a) accuracy and (b) F1 score improvements over the traditional MLP-RF 
		ensemble in the feature space. SECRET uses MLP as the feature space classifier and RF/MLP as the 
		semantic space regressor. Black arrows indicate when the RF regressor is used, wheras the dark blue and 
		dashed arrows correspond to the MLP regressor.}
	\label{fs_fs_mlp_rfc_secret}
	\vspace{-0.4cm}
\end{figure*}

\begin{figure*}[t!]
	\centering
	\subfloat[]{\includegraphics[width=6.0in]{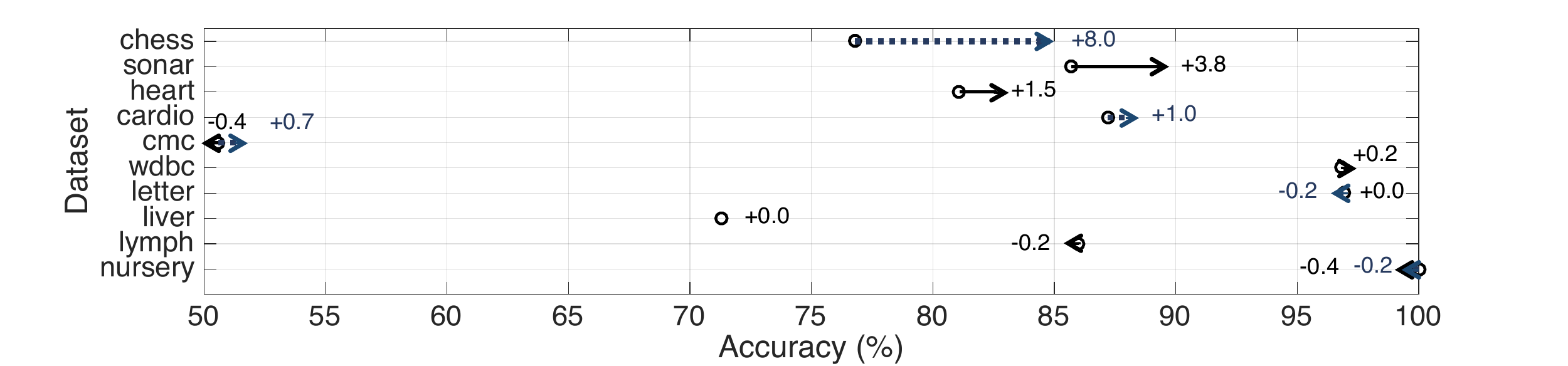}%
		\label{fs_fs_rfc_mlp_secret_acc}}
	\vspace{-0.4cm}
	\hfil
	\subfloat[]{\includegraphics[width=6.0in]{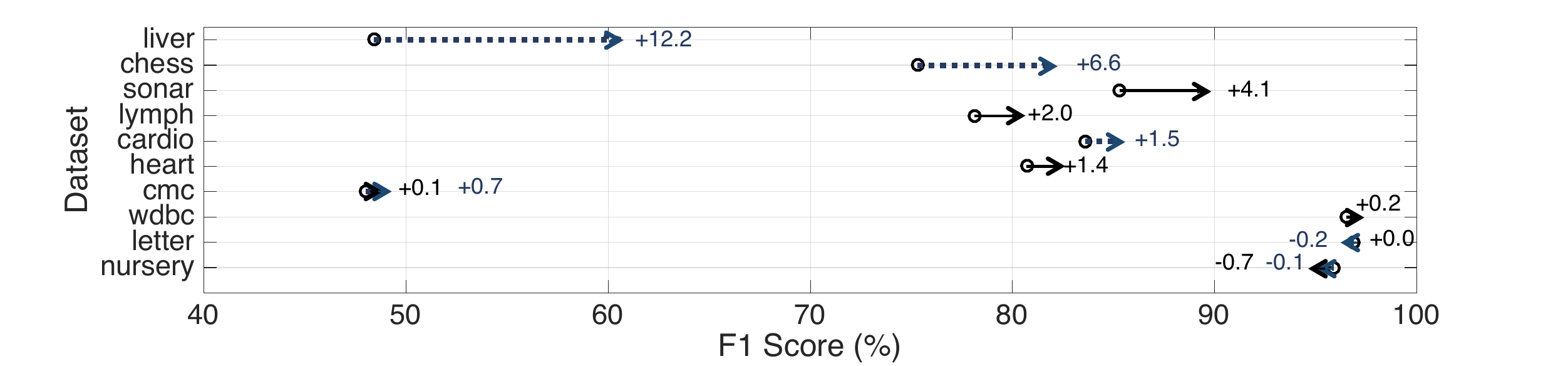}%
		\label{fs_fs_rfc_mlp_secret_f1}}
	\vspace{-0.2cm}
	\hfil
	\caption{SECRET's (a) accuracy and (b) F1 score improvements over the traditional 
		RF-MLP ensemble in the feature space. SECRET uses MLP as the feature space classifier and
		MLP/RF as the semantic space regressor. Black arrows indicate when the MLP regressor is used, 
		whereas the dark blue and dashed arrows correspond to the RF regressor.}
	\label{fs_fs_rfc_mlp_secret}
	\vspace{-0.4cm}
\end{figure*}

Fig.~\ref{fs_fs_mlp_rfc_secret} shows individual and relative classification performance of 
the MLP-RF ensemble and SECRET. Again, the lymph dataset shows performance instability due to its 
size. For chess, cmc, liver, sonar, and cardio datasets, SECRET improves classification performance, 
whereas for the rest (excluding lymph), SECRET either obtains the same or less than 0.8\% lower performance relative to 
the ensemble method. While SECRET improves the accuracy by 0.2 to 7.6\% for the five datasets (excluding lymph), the 
ensemble method only outperforms SECRET by a maximum of 0.8\% in the
case of two datasets. SECRET and the ensemble method obtain comparable 
performance for the nursery and heart datasets.

Fig.~\ref{fs_fs_rfc_mlp_secret} presents accuracy and F1 scores of the RF-MLP ensemble and SECRET. 
If we had not implemented SECRET on top of a traditional supervised (feature space) classifier, but 
built it from ground up, the MLP-RF ensemble would yield the same results as RF-MLP.  However, since 
we would like to compare SECRET with the traditional approach, the hyperparameter values 
are determined by also taking into account the assigned hyperparameter values of the feature space 
block. Since SECRET determines the semantic space hyperparameters using 
joint information from the two spaces, for a fair comparison, we provide the same opportunity to the 
ensemble method while determining the hyperparameter values of the second feature space block. 
Therefore, while RF hyperparameter values take advantage of the knowledge of MLP hyperparameter 
values in Fig.~\ref{fs_fs_mlp_rfc_secret}, MLP hyperparameter values take advantage of the knowledge 
of RF hyperparamenter values in Fig.~\ref{fs_fs_rfc_mlp_secret}. Due to very similar validation set performance ($\leq 1\%$ difference), we were not able 
to determine the regressor type for the nursery and cmc datasets. Therefore, 
we present both results for SECRET with RF and MLP regressors. Although the accuracy improvement
of SECRET on the cmc dataset is inconclusive, SECRET obtains consistent F1 score improvements with 
both regressors. The lymph dataset shows very slight decrease ($0.2\%$) in accuracy; however, 
improvement ($2.0\%$) in F1 score with SECRET. For the letter and nursery datasets, the ensemble 
method has less than 0.1\% to 0.7\% higher F1 score, whereas SECRET has a 0.1 to 12.2\% 
F1 score improvement on the remaining eight datasets.  

\begin{figure*}[t!]
	\centering
	\subfloat[]{\includegraphics[width=6.0in]{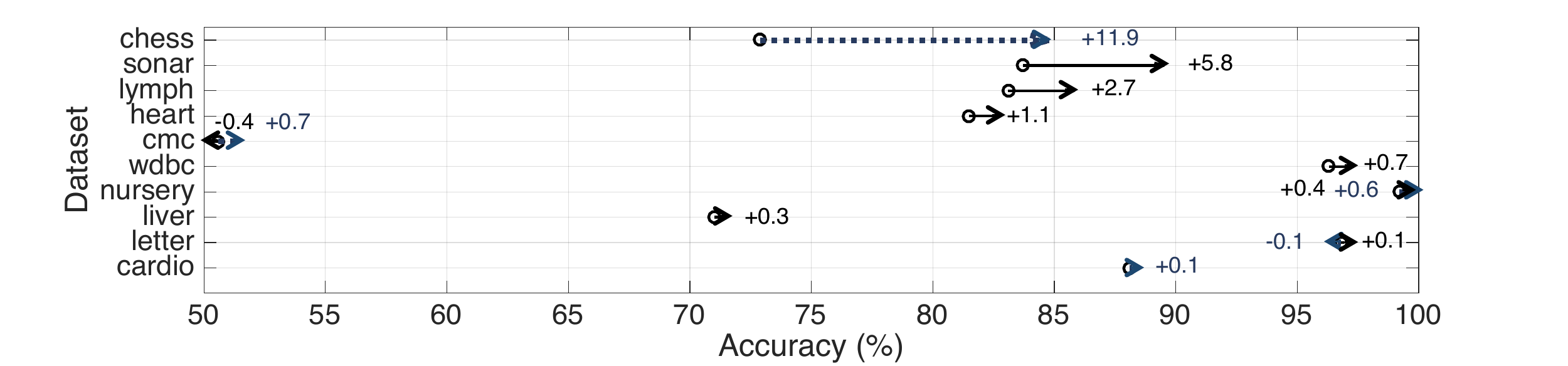}%
		\label{fs_fs_rfc_rfc_secret_acc}}
	\vspace{-0.4cm}
	\hfil
	\subfloat[]{\includegraphics[width=6.0in]{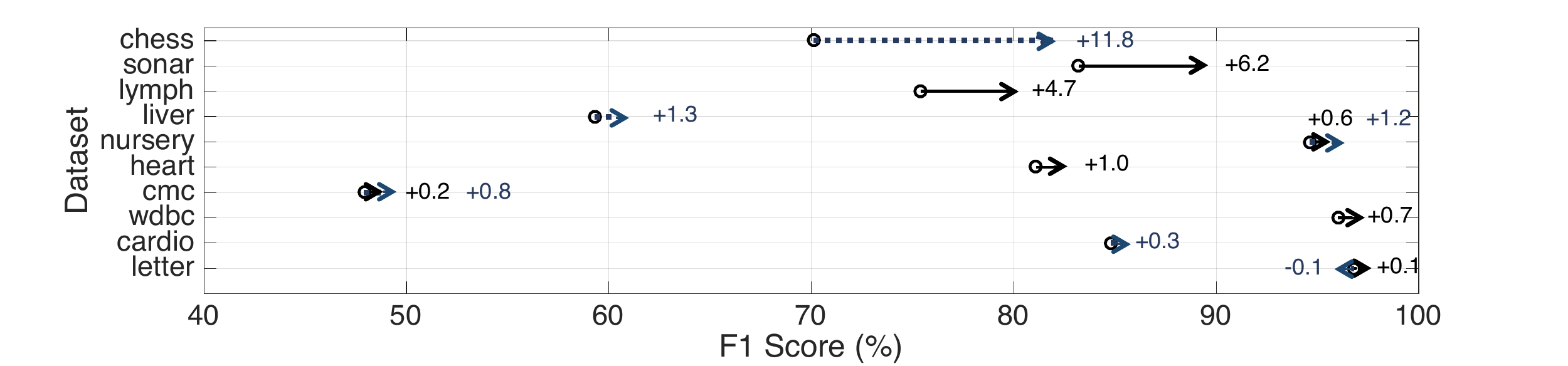}%
		\label{fs_fs_rfc_rfc_secret_f1}}
	\vspace{-0.2cm}
	\hfil
	
	\caption{SECRET's (a) accuracy and (b) F1 score improvements over the traditional RF-RF 
		ensemble in the feature space. SECRET uses MLP as the feature space classifier and MLP/RF as the 
		semantic space regressor. Black arrows indicate when the MLP regressor is used, whereas the dark blue and 
		dashed arrows correspond to the RF regressor.}
	\label{fs_fs_rfc_rfc_secret}
	\vspace{-0.4cm}
\end{figure*}

Fig.~\ref{fs_fs_rfc_rfc_secret} shows the experimental results for the RF-RF ensemble and SECRET. 
For the letter dataset, SECRET obtains comparable performance with the ensemble method. For the 
remaining nine datasets, SECRET provides up to 11.9\% and 11.8\% accuracy and F1 score improvements, 
respectively.

From the above experiments, we can conclude that SECRET leads to either significantly 
higher or comparable classification performance with respect to the ensemble method.

\subsection{RF Decision Node Depth} \label{rf_node_depth}

In this section, we provide insight into 
how SECRET's semantic space RF models differ from the traditional feature space ones. Since SECRET 
uses meaning-based relationships among labels, it is able to divide the classes into 
`easy-to-classify' and `difficult-to-classify'. We expect SECRET to adjust the RF decision node 
depths according to both the semantic relationships among labels and data characteristics, and 
traditional approaches to adjust only according to data characteristics. Therefore, we hypothesize 
that the decision node depth for different classes varies more in SECRET compared to the traditional 
approaches as labels show heterogeneous distribution in the semantic space and SECRET is able to 
integrate this information into the classification task (focus on `difficult-to-classify' classes 
in deeper nodes with the help of its semantic space component).

We carry out the RF decision node depth experiments on six datasets (cmc, chess, lymph, cardio, 
nursery, and letter) to validate our hypothesis. The remaining four datasets (sonar, liver, wdbc, 
and heart) 
have two classes. When one class is assigned, the other class also gets distinguished. Therefore, 
the variance of the decision node depth for these four datasets tends to zero, which is 
not informative. For the six datasets that include three or more classes, we take each decision tree 
in the RF model and assess the decision nodes, their depth, and assigned classes. It is important to 
note that the RF models for the feature and semantic spaces are obtained with a classifier and 
a regressor, respectively. To make a fair comparison, we convert the semantic-space RF regressor to 
a classifier by performing labeling with only the regressor outputs. For both models, within a tree, 
we compute the average decision node depth for each class. We repeat this process for each tree to 
assess the overall average decision node depth for the RF model. 

\begin{table*}[t!]
	\caption{Average RF Decision Node Depth, Overall RF Node Depth Variance and Classification Performance on the cmc Dataset}
	\centering
	\begin{tabular}{l |c c c |c |c c}
		
		\hline
		\multicolumn{1}{l|}{Approach}& \multicolumn{3}{c|}{Average RF Node Depth}& \multicolumn{1}{c|}{Overall Variance}& \multicolumn{2}{c}{Classification Performance}\\[0.5ex]
		& no use & long-term methods & short-term methods&  of RF Node Depth& Accuracy (\%) & F1 Score (\%) \\[0.5ex]
		\hline\hline
		Traditional Classifier&11.6&12.4&12.4&0.2&49.8&46.8\\
		Traditional Ensemble&11.6&12.5&12.5&0.2&51.3&48.5\\
		(built on top of MLP)&&&&&&\\
		Traditional Ensemble&11.5&12.4&12.4&0.2&50.6&47.9\\
		(built on top of RF)&&&&&&\\

		\bf{SECRET}&\bf{11.0}&\bf{12.3}&\bf{12.2}&\bf{0.4}&\bf{52.9}&\bf{50.5}\\
		\bf{(built on top of MLP)}&&&&&&\\
		\bf{SECRET}&\bf{11.0}&\bf{12.3}&\bf{12.2}&\bf{0.4}&\bf{51.3}&\bf{48.7}\\
		\bf{(built on top of RF)}&&&&&&\\
		\hline	
		\hline
	\end{tabular}
	\label{rf_depth_table}
	\vspace{-0.4cm}
\end{table*}

\begin{figure}[t!]
	\centering
	\vspace{-0.3cm}
	\subfloat[]{\includegraphics[width=3.0in]{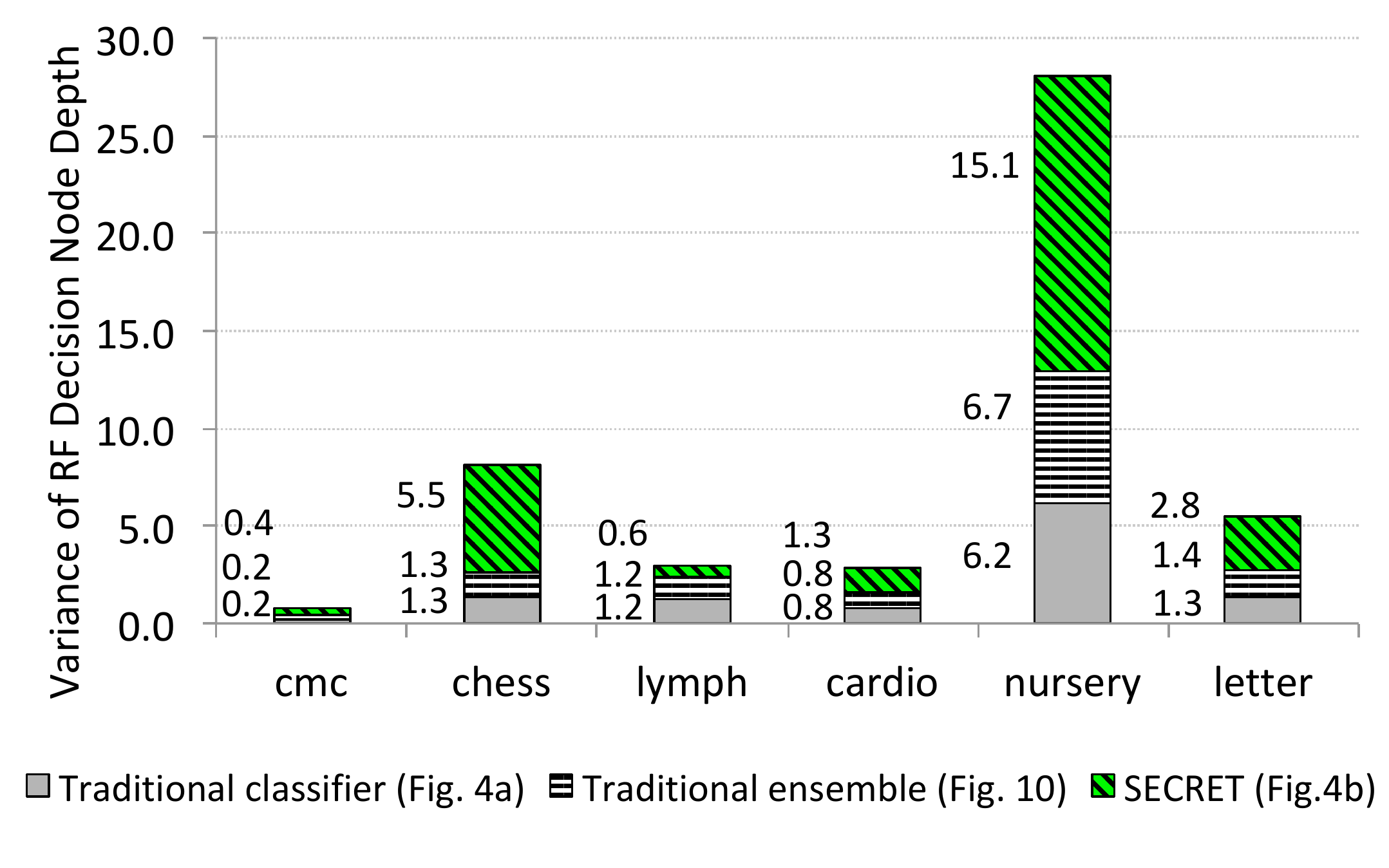}%
		\label{rf_depth_mlp_rf}}
	\vspace{-0.4cm}
	\hfil
	\subfloat[]{\includegraphics[width=3.0in]{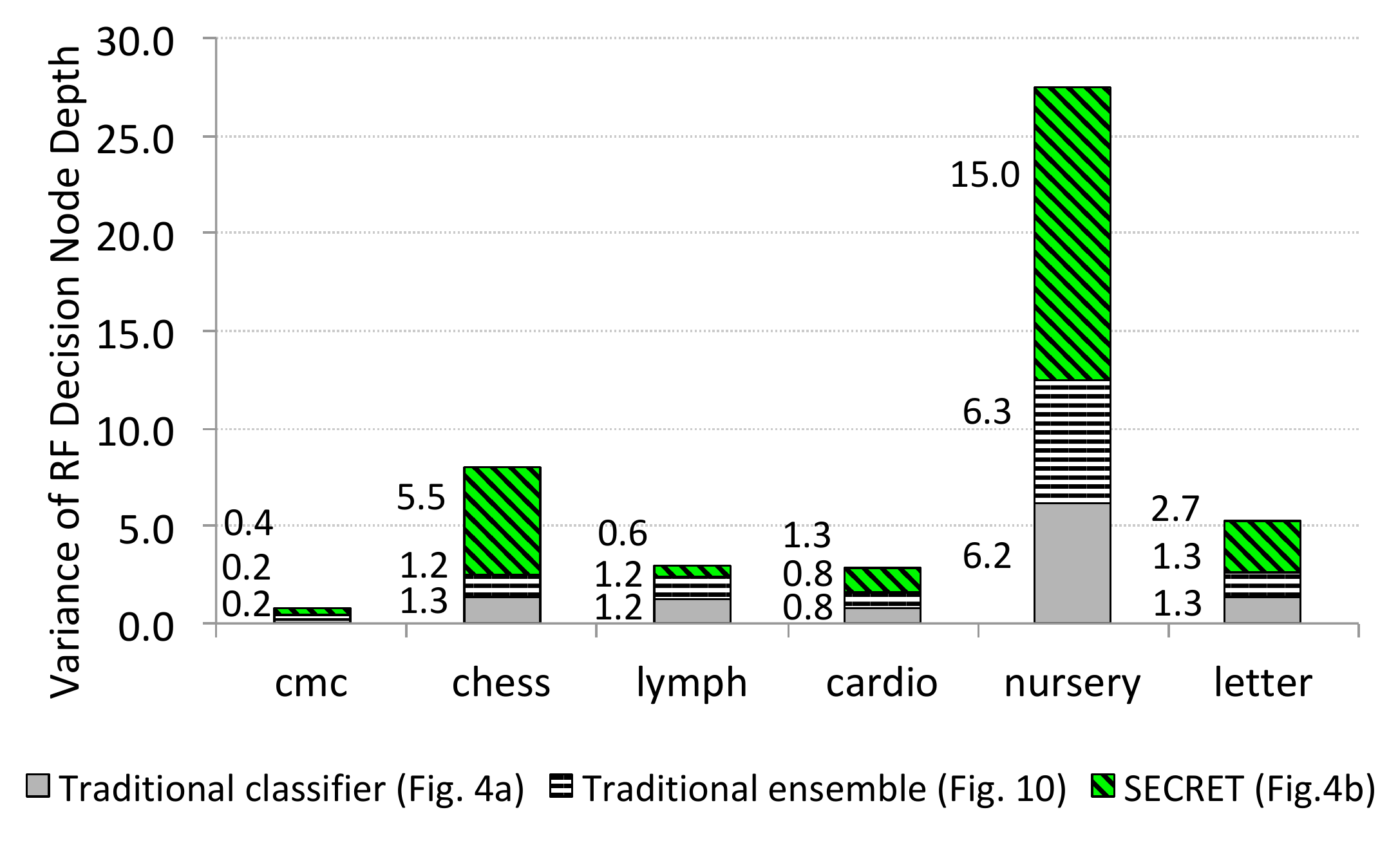}%
		\label{rf_depth_rf_rf}}
	\vspace{-0.2cm}
	\hfil
	\caption{Variance of RF decision node depth for traditional RF classifier and traditional ensemble and SECRET built on top of (a) MLP and (b) RF.}
	\label{rf_depth}
	\vspace{-0.5cm}
\end{figure}

As an example, Table~\ref{rf_depth_table} shows RF decision node depth variance and classification 
performance on the cmc dataset for both the traditional approaches and SECRET. While the traditional 
approaches assign `no use,' `long-term methods,' and `short-term methods' at closer node depths by 
taking data characteristics into account, SECRET uses both the data characteristics and semantic 
relationships among labels (Fig.~\ref{cmc_label_position_ss}). As the `no use' class is located 
farther away (in Euclidean distance) from the `long-term methods,' and `short-term methods' classes, 
SECRET assigns `no use' to shallower depths and focuses on details to distinguish `long-term 
methods,' and `short-term methods' at deeper nodes. We summarize this result into one value,
which is the overall variance of the RF node depth among all classes and 10 folds. While SECRET
takes the heterogeneous distribution of the labels (in the semantic space) into account and
shapes the tree depths accordingly, the traditional classifiers or ensembles ignore this point. They 
tend to assign labels at similar average node depths, resulting in a smaller overall variance 
compared to SECRET. As a result of SECRET's directed attention to `easy-to-classify' and 
`difficult-to-classify' classes, it outperforms the traditional approaches, as shown in the right column of Table~\ref{rf_depth_table}. For the remaining datasets, we carried 
out the same analyses. Fig.~\ref{rf_depth} shows the overall variance of RF decision node 
depth for `Traditional classifier,' `Traditional ensemble,' and `SECRET.' In 
Fig.~\ref{rf_depth_mlp_rf} and Fig.~\ref{rf_depth_rf_rf}, `Traditional classifier' represents 
the variance of RF model's decision node depth. In Fig.~\ref{rf_depth_mlp_rf}, 'Traditional Ensemble' 
and `SECRET' represent variance of decision node depth of RF models that are built on top of 
the MLP model, as shown in Fig.~\ref{secret_block} and Fig.~\ref{ensemble_block}, respectively. 
In Fig.~\ref{rf_depth_rf_rf}, 'Traditional Ensemble' and 
`SECRET' are built on top of an RF model. In five of the datasets (except lymph), we observe a larger 
variance in the overall decision node depth of SECRET compared to the traditional approaches. In 
line with this observation, SECRET obtains up to 11.2\% and 12.1\% accuracy and F1 
score improvements, respectively, over the traditional classifier and up to 
7.6\% and 7.4\% accuracy and F1 score improvement in accuracy and F1 score over the 
traditional ensemble method depicted in Fig.~\ref{ensemble_block}. For the other 
case shown in Fig.~\ref{rf_depth_rf_rf}, SECRET obtains up to $11.8\%$ and $11.7\%$ 
accuracy and F1 score improvements, respectively, over the traditional classifier and 
up to $11.9\%$ and $11.8\%$ improvements in accuracy and F1 score, respectively, over the traditional 
ensemble method. For the letter dataset, we observe comparable performance 
(maximum $0.3\%$ decrease in accuracy/F1 score) with the traditional approaches. For the lymph 
dataset, while RF node depth variance is smaller for SECRET, we observe $0.1\%$ to $3.6\%$ accuracy and $0.0\%$ to $5.8\%$
improvement over the traditional approaches. This is inconclusive. As we also obtain inconclusive 
results throughout Section~\ref{results} due to its size, we do not discuss the lymph dataset
further. 

Overall, a larger variance in RF node depth indicates that SECRET is distinguishing 
`easy-to-classify' and `difficult-to-classify' cases more clearly than the traditional approaches and 
focusing on detailed characteristics at deeper nodes to separate the `difficult-to-classify' cases 
further. As a result, we observe an enhancement in classification performance with SECRET. This 
is commensurate with our hypothesis.

\section{Related Work} \label{related_work}

Next, we focus on related studies in both the feature and semantic spaces.

Enhancing classification performance of the ML algorithms has been a well-targeted area of research 
for decades. Various approaches have been proposed. These include data augmentation 
\cite{van2001art}, data generation \cite{he2008adasyn}, boosting \cite{freund1999short}, ensemble 
learning \cite{dietterich2002ensemble}, and dimensionality reduction \cite{van2009dimensionality}. In 
addition to these promising techniques, various ML algorithms (information-based, similarity-based, 
probability-based, and error-based \cite{kelleher2015fundamentals}) and architectures have been 
designed. Specifically, for big data, neural network models \cite{alom2019state} have revolutionized 
the classification task due to their ability to model complex data-label relationships. Although 
these algorithms and techniques have made significant contributions to enhancing classification 
performance, they all operate in the feature space.  SECRET closes the gap between the feature and 
semantic spaces. 

If we change our perspective and look at the related work in the semantic space, we observe that 
word representations have been widely used in NLP applications. Liu et al.~\cite{liu2018task} 
proposed a novel task-oriented word embedding method to assess the salient word for the text 
classification task. All analyses are carried out in the semantic space. Kusner et 
al.~\cite{kusner2015word} introduced a novel distance metric (Word Mover's Distance) to effectively 
model the text documents with a set of word vectors. Vector representations act as features in 
a traditonal classification task and are mapped to a pre-defined set of labels with 
the k-Nearest Neighbor algorithm. This is a feature space approach since word vectors are used as 
features and mapped to a specific set of labels, without considering the meaning-based relationships 
among labels. Bordes et al.~\cite{bordes2014open} targeted question answering by representing the 
question in a vector form in the semantic space and mapping it to the answer again in the semantic 
space. Bengio and Heigold \cite{bengio2014word} go far away from the semantic space 
by training vector representations of words without considering their meaning 
relationships, but targeting how similar the words sound. 
Vectors of sound-alike (not semantically similar) words have a smaller Euclidean 
distance between them. Wang et al. \cite{wang2016cnn} focused on the design of a CNN-RNN framework 
that maps image data to label embedding to perform multi-label classification while taking 
label co-occurrence and semantic redundancy into account. The proposed approach targeted only the 
image datasets with a fixed ML design. Palatucci et al.~\cite{palatucci2009zero} and Socher 
et al.~\cite{socher2013zero} carried out zero-shot learning by mapping real-world data to semantic 
vector representations of words. Karpathy and Fei-Fei \cite{karpathy2015deep} obtained figure 
captions using image datasets and word embeddings. The approaches presented in 
\cite{palatucci2009zero}, \cite{socher2013zero}, and \cite{karpathy2015deep} are limited to the 
semantic space. They only targeted correlations between data features and semantic relationships. 

Overall, the above-mentioned approaches have had a significant influence on the development of NLP 
applications; however, they exploit either the feature space or the semantic space when performing 
classification. SECRET integrates these two spaces. Thus, SECRET can be differentiated from previous 
work and looks at real-world classification tasks in a new way.

\begin{figure}[t!]
	\centering
	\subfloat[]{\includegraphics[width=2.7in]{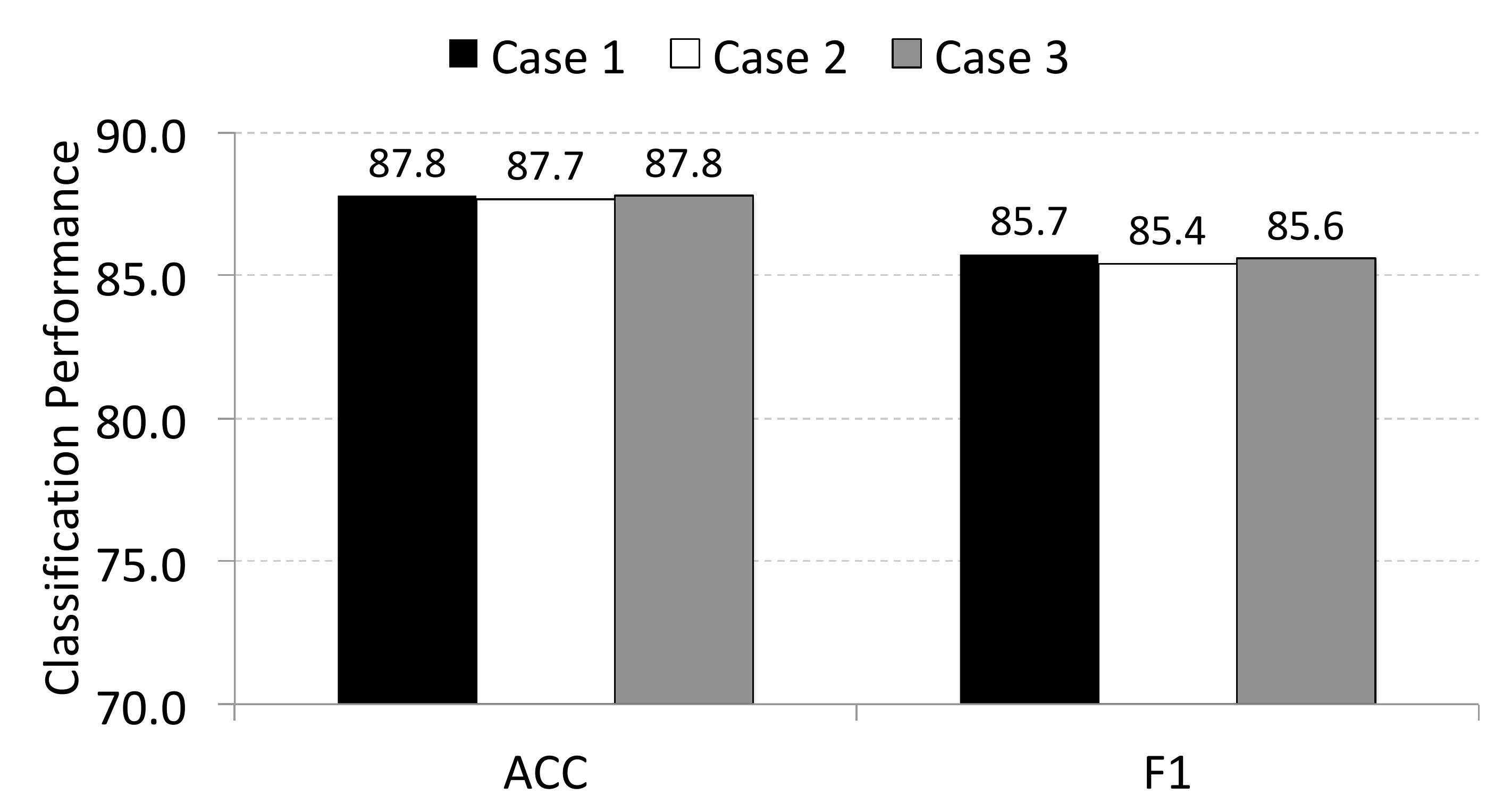}%
	\vspace{-0.6cm}
		\label{discussion_synonym_mlp_rfc}}	
	\vspace{-0.4cm}
	\hfil	
	\subfloat[]{\includegraphics[width=2.7in]{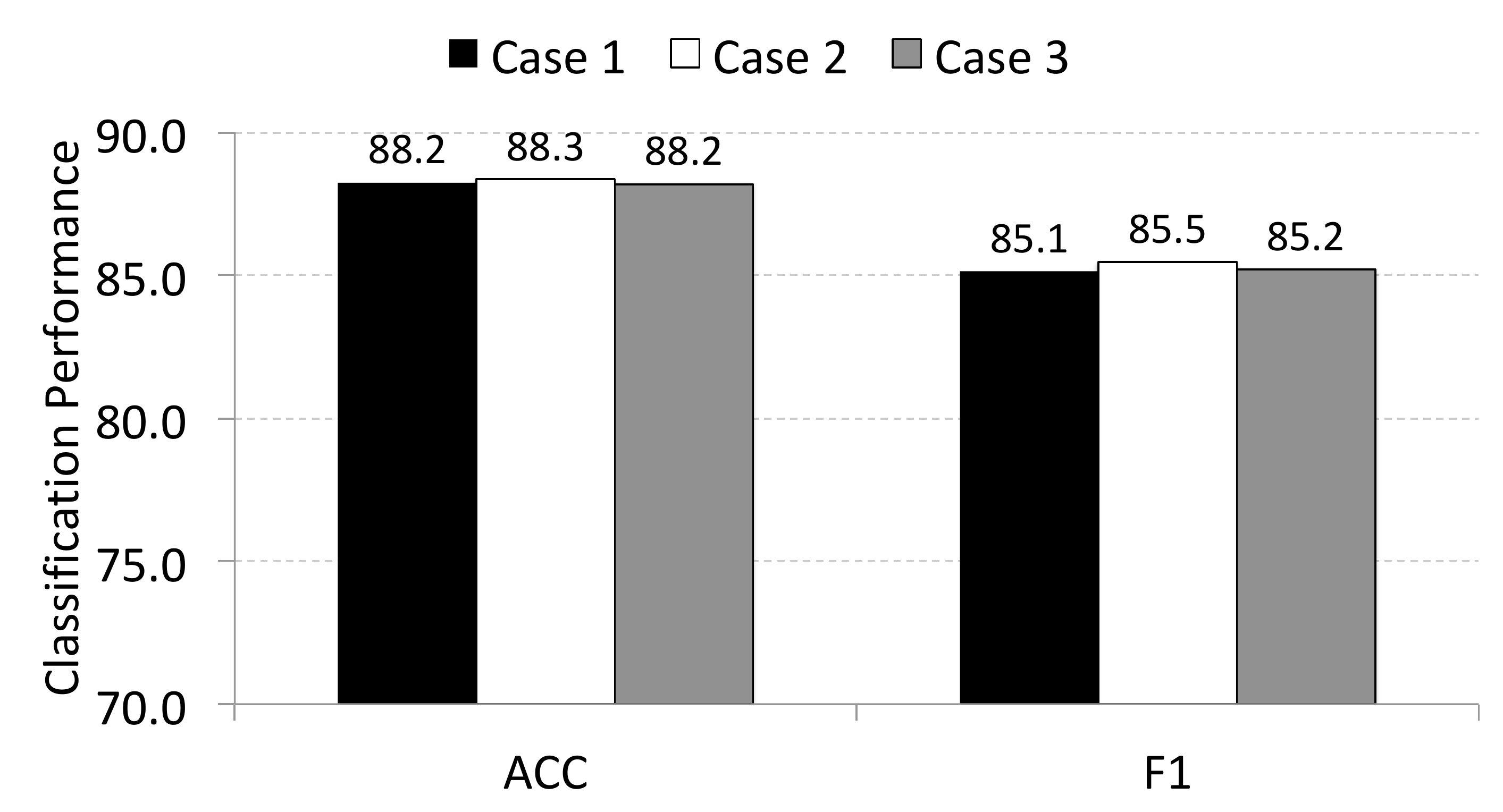}%
	\vspace{-0.6cm}
		\label{discussion_synonym_rfc_rfc}}
	\vspace{-0.2cm}
	\hfil
	\caption{Classification performance of SECRET on the cardio dataset when built with 
(a) MLP and (b) RF as the feature space classifier and RF as the semantic space regressor. Case 
1 represents the performance of the cardio dataset when the labels are used as they are. Case 2 
and Case 3 represent the performance when the `REM sleep' label is replaced with `paradoxical sleep' 
and `dreaming sleep,' respectively.}
	\label{discussion_synonym}
	\vspace{-0.4cm}
\end{figure}

\section{Discussion} \label{discussion}
In this section, we analyze SECRET from different perspectives.
\subsection{What to do for classes with the same label, but different meanings?}
Word embedding algorithms, such as GloVe, capture the word's semantic relationships with the 
remaining words in the dictionary. Words with close meanings are represented by closely-spaced 
vectors in the semantic space. Therefore, vectors do not represent the sounding, but the meaning 
relationships between words. None of the datasets used in our experiments (Table \ref{dataset}) 
included homographs (words with the same spelling, but different meanings). However, in case of a 
homograph (such as `spring'), the corresponding semantic vectors should be obtained with 
context-aware word embedding algorithms, as proposed by studies in \cite{liu2015learning}, 
\cite{liu2015topical}, \cite{nguyen2017overview}, \cite{lee2017muse}.

\subsection{How to decide on the semantic vector of a class whose label has synonyms?}
We hypothesize that SECRET provides very similar classification outputs when the synonyms of a class 
label are used interchangeably. We tested this hypothesis on the cardio dataset.  As indicated by 
studies in \cite{jasper1971acetylcholine} and \cite{greenberg1972effect}, the `REM sleep' label means 
the same as `paradoxical sleep' and `dreaming sleep.' Fig.~\ref{discussion_synonym} shows the 
classification accuracy and F1 scores for three cases when SECRET is built on top of the MLP and 
RF feature space classifiers. In all cases, SECRET obtains nearly the same classification 
performance, buttressing our hypothesis.

\begin{figure}[t!]
	\centering
	\subfloat[]{\includegraphics[width=2.3in]{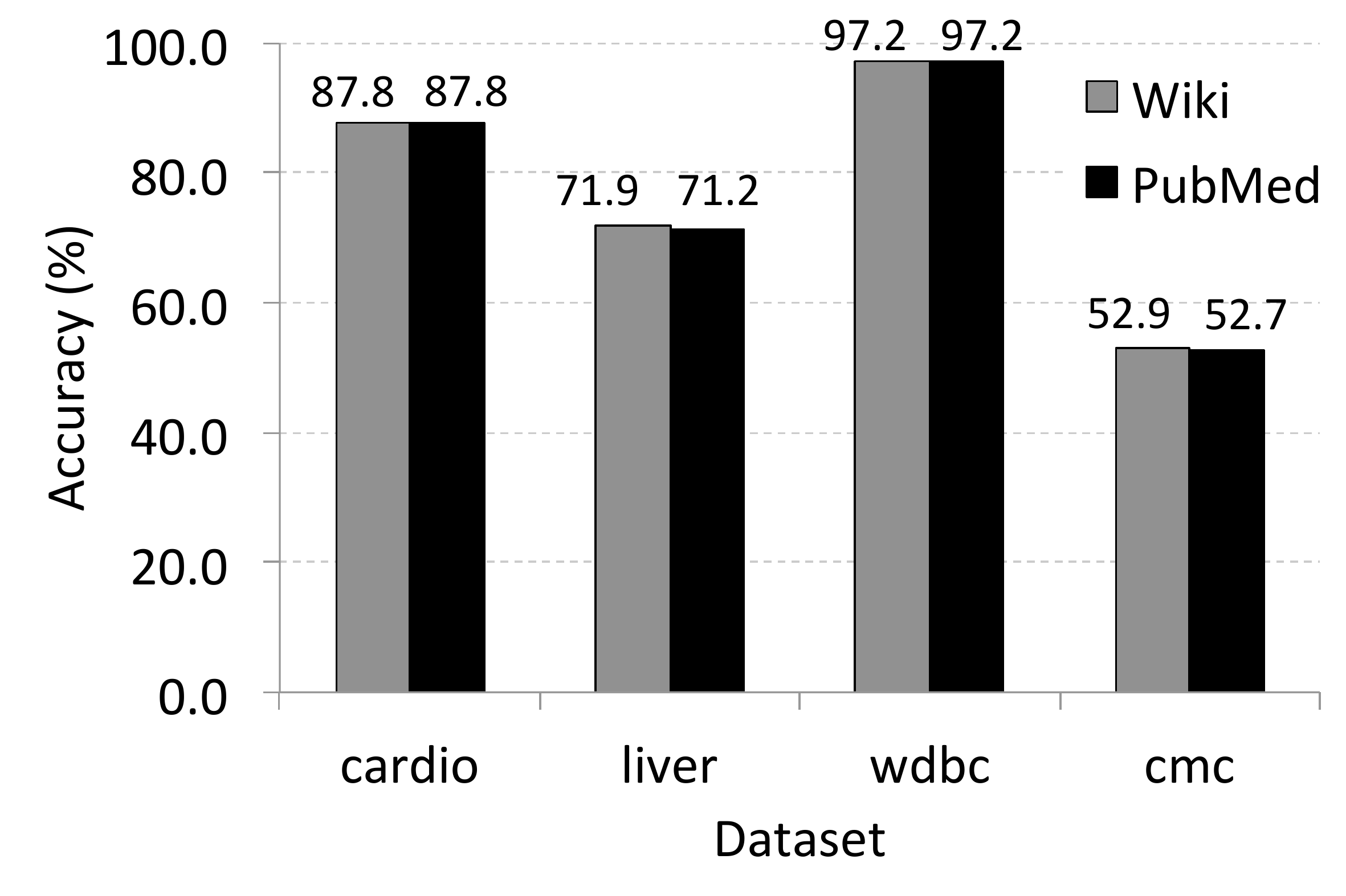}%
		\vspace{-1.2cm}
		\label{discussion_pubmed_mlp_acc}}	
	\vspace{-0.4cm}
	\hfil	
	\subfloat[]{\includegraphics[width=2.3in]{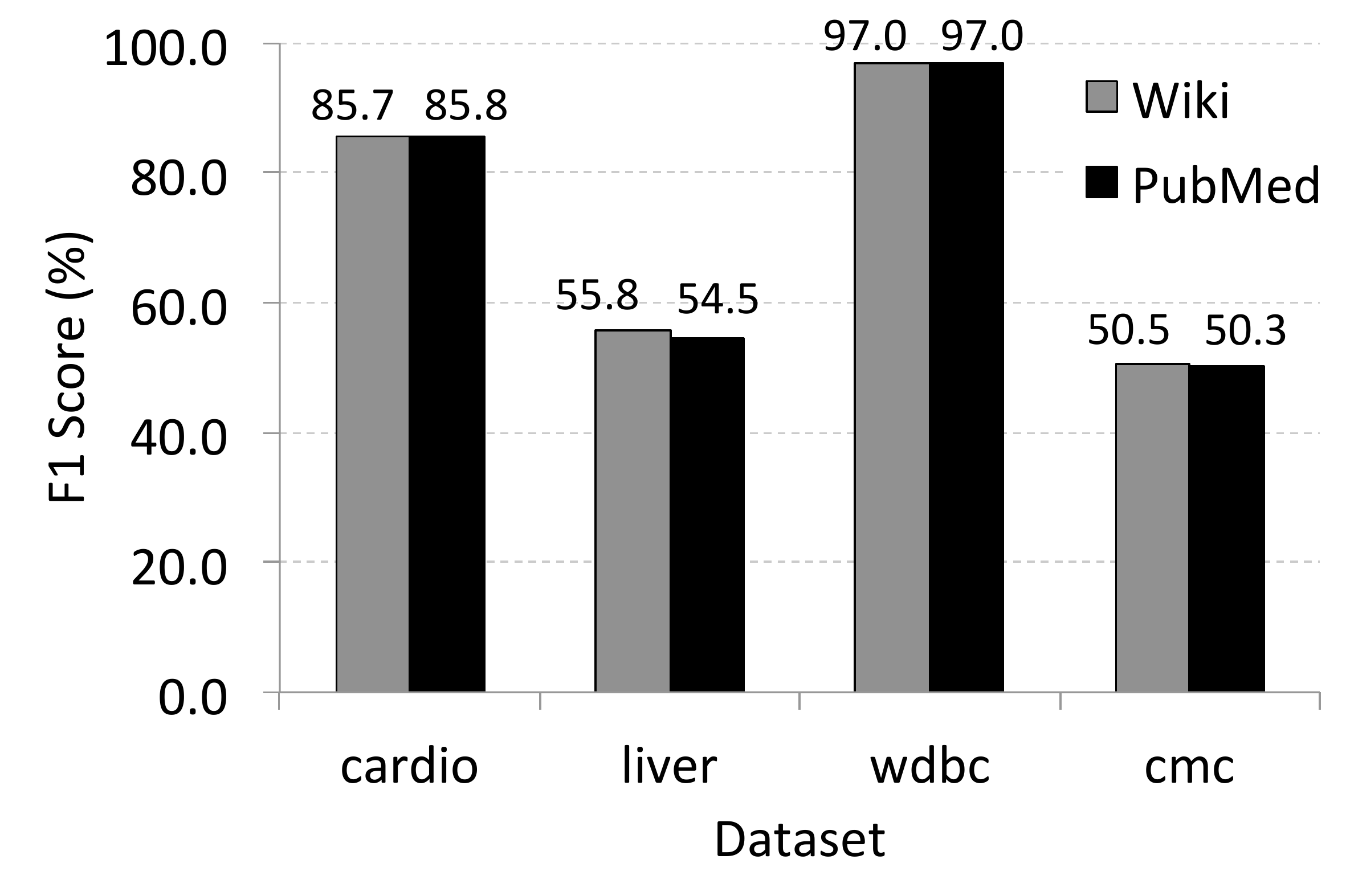}%
		\vspace{-0.4cm}
		\label{discussion_pubmed_mlp_f1}}
	\vspace{-0.2cm}
	\hfil
	\caption{Comparison of SECRET's (a) accuracy and (b) F1 score values when built with 
		pretrained Wikipedia 2014 + Gigaword (Wiki) and PubMed word vectors. SECRET uses MLP as the feature 
		space classifier and RF as the semantic space regressor.}
	\label{discussion_pubmed_mlp}
	\vspace{-0.4cm}
\end{figure}

\subsection{Do the improvements come from the selected word embedding or SECRET? How does SECRET 
perform when a different word embedding is introduced?}
\begin{figure}[t!]
	\centering
	\subfloat[]{\includegraphics[width=2.3in]{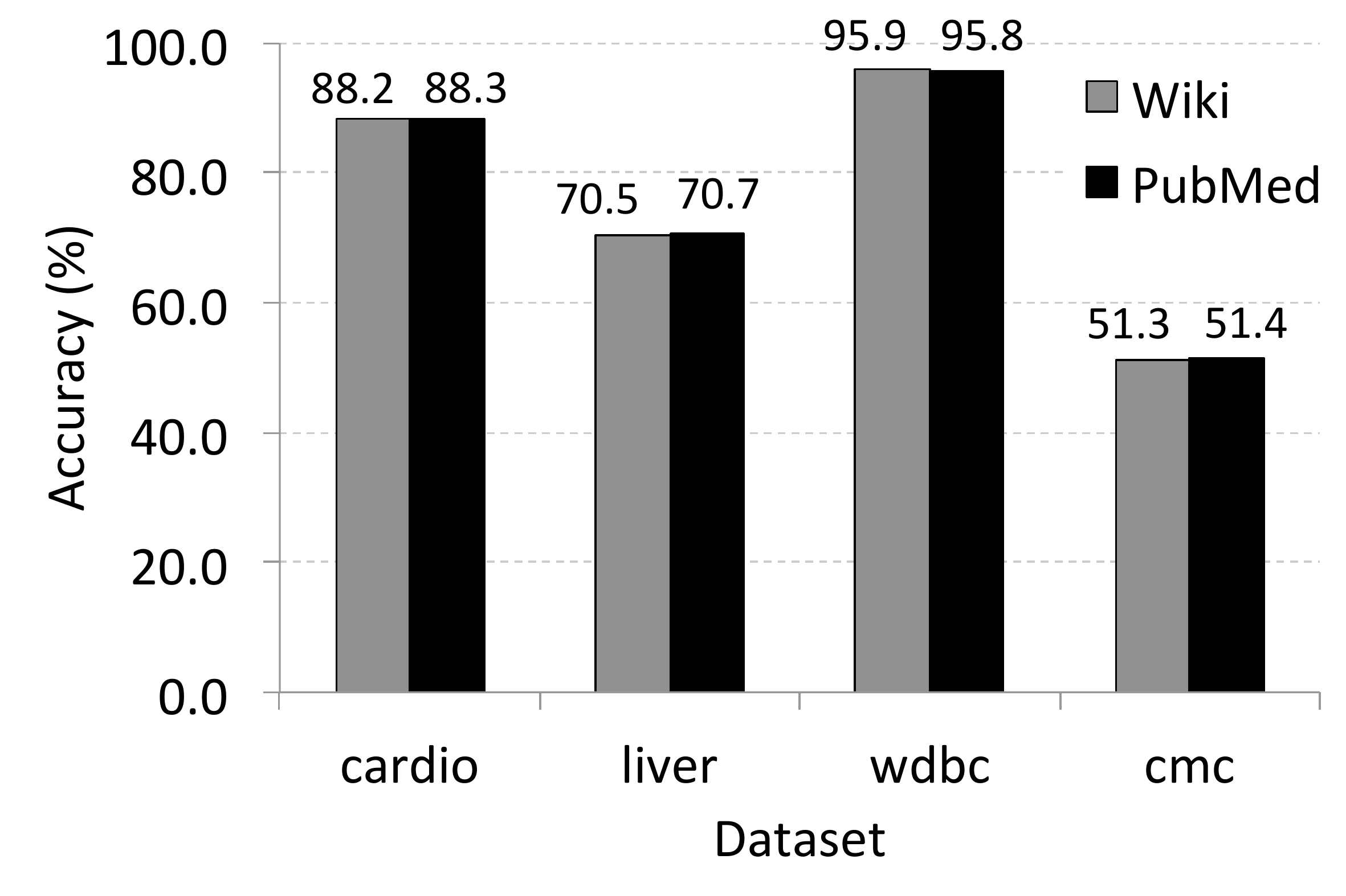}%
		\vspace{-1.0cm}
		\label{discussion_pubmed_rf_acc}}	
	\vspace{-0.3cm}
	\hfil	
	\subfloat[]{\includegraphics[width=2.3in]{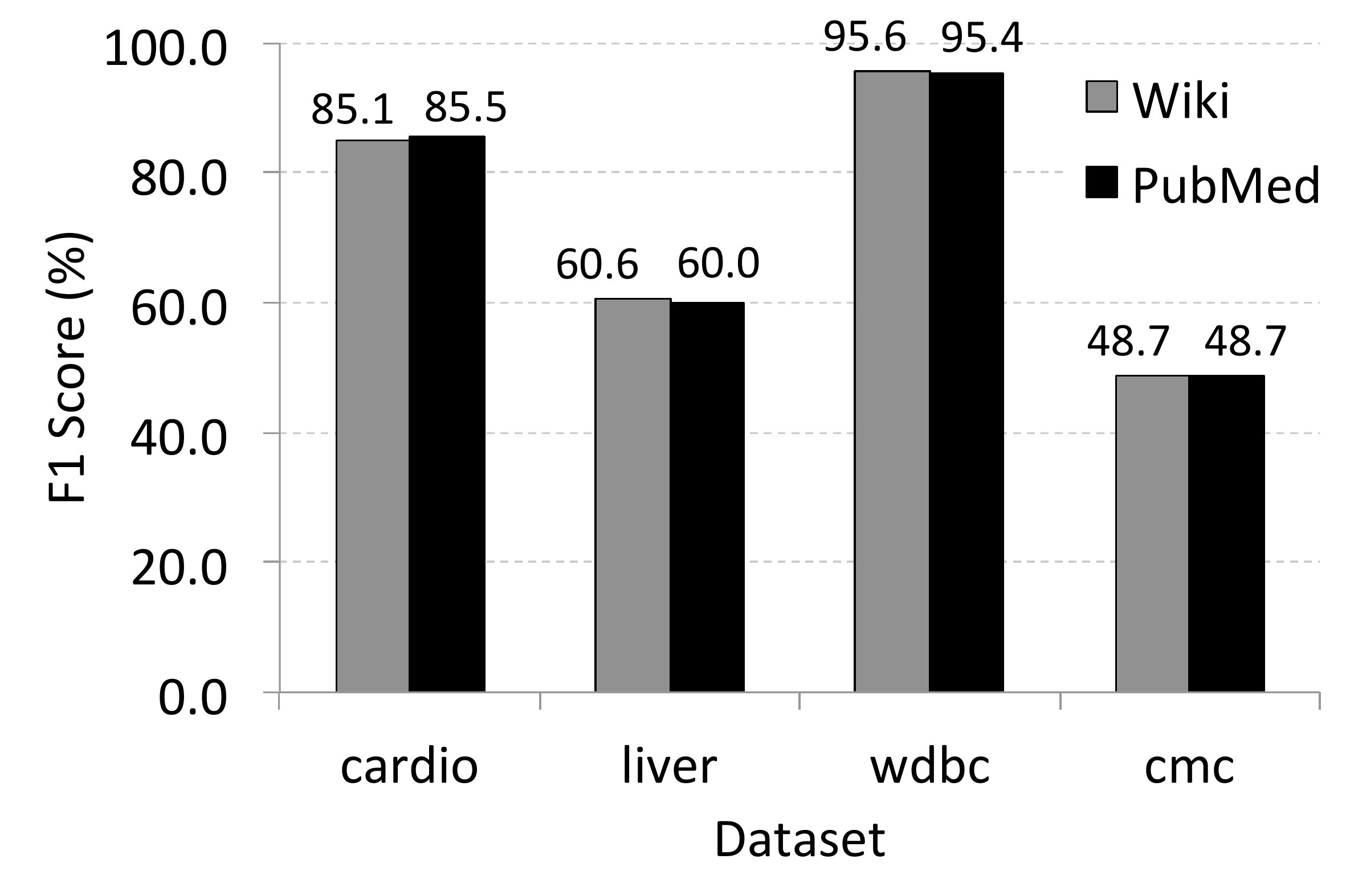}%
		\vspace{-0.4cm}
		\label{discussion_pubmed_rf_f1}}
	\hfil
	\caption{Comparison of SECRET's (a) accuracy and (b) F1 score values when built with 
		pretrained Wikipedia 2014 + Gigaword (Wiki) and PubMed word vectors. SECRET uses RF as the feature 
		space classifier and RF as the semantic space regressor.}
	\label{discussion_pubmed_rf}
	\vspace{-0.4cm}
\end{figure}

In order to demonstrate the effectiveness of SECRET and its performance stability on various word 
vectors, we repeated the experiments with a different set of pretrained word vectors 
\cite{word2vec_pretrained_biomedical} that is obtained using PubMed texts and the word2vec 
algorithm. PubMed stores biomedical literature and the corresponding word vectors are 
200-dimensional. Considering the application area and size of the datasets (we need to avoid 
overfitting), we focus on the cardio, liver, wdbc, and cmc datasets. 
Fig.~\ref{discussion_pubmed_mlp} shows the classification performance of SECRET when built with MLP 
as the feature space classifier and RF as the semantic space regressor. Similarly, 
Fig.~\ref{discussion_pubmed_rf} shows the performance with RF as the feature space classifier
as well as the semantic space regressor. In both figures, we observe nearly identical classification 
performance for Wiki and PubMed-based word vectors. These results provide evidence that the 
classification performance improvements demonstrated in Figs.~\ref{fs_mlp_secret}, 
\ref{fs_rfc_secret}, \ref{fs_fs_mlp_mlp_secret}, \ref{fs_fs_mlp_rfc_secret}, 
\ref{fs_fs_rfc_mlp_secret}, and \ref{fs_fs_rfc_rfc_secret} do not arise from Wiki-based word vectors, 
but the integration of semantic information into the classification task.

\subsection{Does the semantic space regressor by itself perform better than SECRET?}
\begin{figure*}[t!]
	\centering
	\subfloat[]{\includegraphics[width=6.0in]{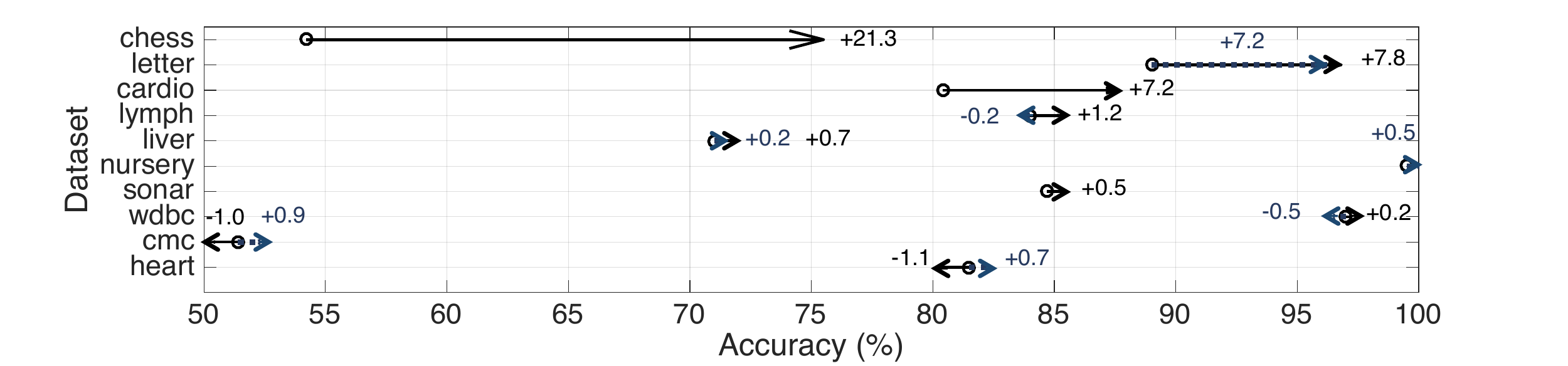}%
		\label{ss_mlp_secret_acc}}
	\vspace{-0.4cm}
	\hfil
	\subfloat[]{\includegraphics[width=6.0in]{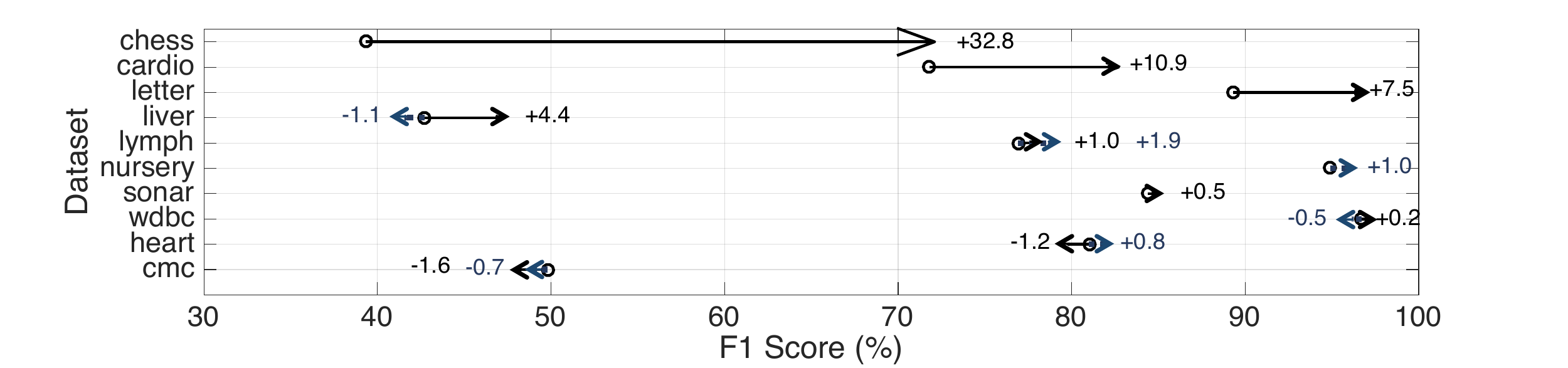}%
		\label{ss_mlp_secret_f1}}
	\vspace{-0.2cm}
	\hfil
	\caption{SECRET's (a) accuracy and (b) F1 score improvements over the MLP regressor in the 
semantic space. SECRET uses MLP as the semantic space regressor and RF/MLP as the feature space 
classifier. Black arrows indicate when the RF classifier is used, whereas the dark blue and dashed arrows 
correspond to the MLP classifier.}
	\vspace{-0.4cm}
	\label{ss_mlp_secret}
\end{figure*}

\begin{figure*}[t!]
	\centering
	\subfloat[]{\includegraphics[width=6.2in]{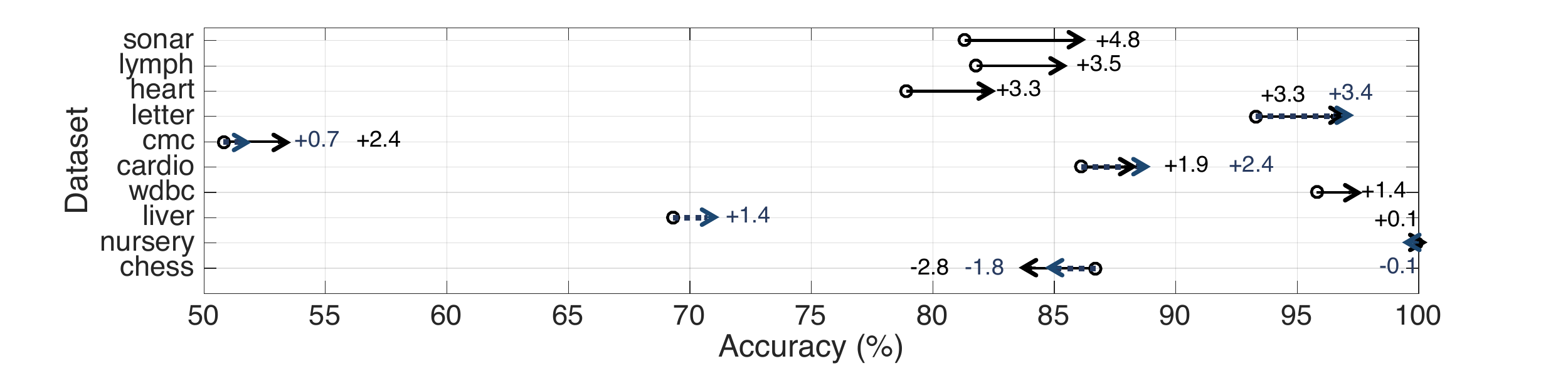}%
		\label{ss_rfc_secret_acc}}
	\vspace{-0.4cm}
	\hfil
	\subfloat[]{\includegraphics[width=6.2in]{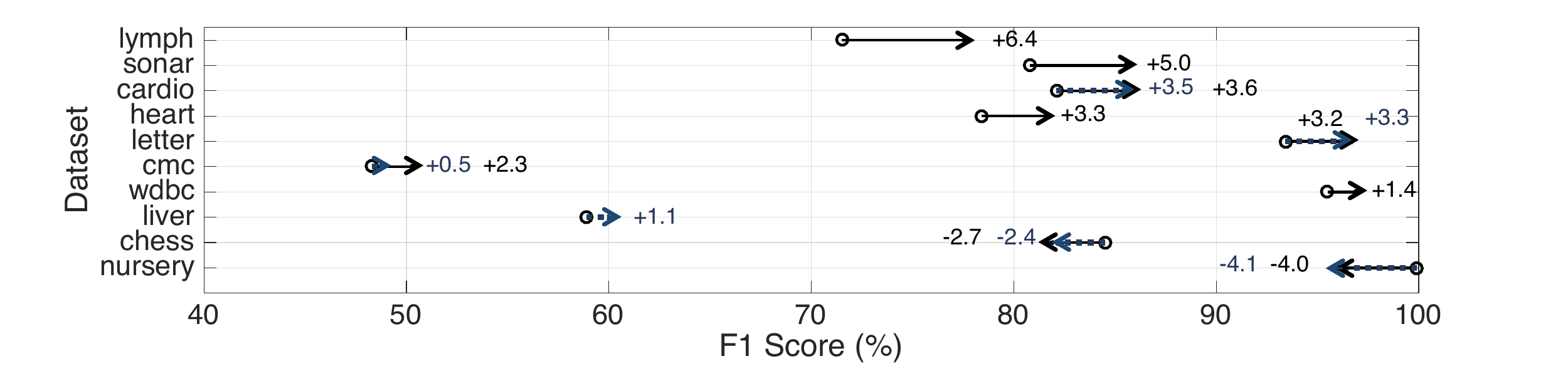}%
		\label{ss_rfc_secret_f1}}
	\vspace{-0.2cm}
	\hfil
	\caption{SECRET's (a) accuracy and (b) F1 score improvements over the RF regressor in the 
semantic space. SECRET uses RF as the semantic space regressor and MLP/RF as the feature space 
classifier. Black arrows indicate when the MLP classifier is used, whereas the dark blue and dashed arrows 
correspond to the RF classifier.}
	\vspace{-0.4cm}
	\label{ss_rfc_secret}
\end{figure*}

We carried out additional experiments for the semantic space (only) case. We show the results in 
Fig.~\ref{ss_mlp_secret} and Fig.~\ref{ss_rfc_secret} for MLP and RF regressors, respectively. As 
in the case of the feature space (only), SECRET dominates the semantic space (only) too. This result 
was expected since SECRET integrates the feature and semantic spaces to benefit from their 
complementary advantages. 

\section{Conclusion and Future Work} \label{conclusion}
In this article, we introduced a new dimension (semantic space) to
the feature space based decision-making employed in ML algorithms and encapsulated it
in a dual-space classification approach called SECRET. As opposed to 
traditional approaches, SECRET maps data to labels while integrating
meaning-based relationships among labels. We analyzed SECRET's
classification performance on ten datasets that represent
different real-world applications. Among ten different real-world datasets, compared to traditional supervised
learning, SECRET achieved up to 14.0\% accuracy and 13.1\% F1
score improvements. Compared to ensemble methods, SECRET
achieved up to 12.7\% accuracy and 13.3\% F1 score improvements. We also took a step toward understanding how SECRET builds the semantic space component and its 
impact on overall classification performance. We posit that, in future work, further improvements in
SECRET's overall classification performance and feature/semantic space characteristics 
can be made as follows. First, further analyses of different datasets are needed to support 
extensive applicability of SECRET. Second, although MLP and RF are well-known
supervised ML algorithms, other ML algorithms need to be analyzed in this context. 
Third, semantic vectors could be trained specially for SECRET and the corresponding application 
of interest, as done in the case of intrinsic and extrinsic analyses in NLP 
\cite{resnik2010evaluation}, \cite{zhai2016intrinsic}. Fourth, detailed classification analyses
need to be carried out for the multilabel classification task, where SECRET can be implemented with 
multilabel classification and regression algorithms targeted at the feature and 
semantic spaces, respectively. Finally, in addition to the feature and 
semantic spaces, other information sources for classification should be explored.

\ifCLASSOPTIONcompsoc

\ifCLASSOPTIONcaptionsoff
  \newpage
\fi

\bibliographystyle{IEEEtran}
\bibliography{references}

\begin{thebibliography}{10}
\providecommand{\url}[1]{#1}
\csname url@samestyle\endcsname
\providecommand{\newblock}{\relax}
\providecommand{\bibinfo}[2]{#2}
\providecommand{\BIBentrySTDinterwordspacing}{\spaceskip=0pt\relax}
\providecommand{\BIBentryALTinterwordstretchfactor}{4}
\providecommand{\BIBentryALTinterwordspacing}{\spaceskip=\fontdimen2\font plus
\BIBentryALTinterwordstretchfactor\fontdimen3\font minus
  \fontdimen4\font\relax}
\providecommand{\BIBforeignlanguage}[2]{{%
\expandafter\ifx\csname l@#1\endcsname\relax
\typeout{** WARNING: IEEEtran.bst: No hyphenation pattern has been}%
\typeout{** loaded for the language `#1'. Using the pattern for}%
\typeout{** the default language instead.}%
\else
\language=\csname l@#1\endcsname
\fi
#2}}
\providecommand{\BIBdecl}{\relax}
\BIBdecl

\bibitem{hirschberg2015advances}
J.~Hirschberg and C.~D. Manning, ``Advances in natural language processing,''
  \emph{Science}, vol. 349, no. 6245, pp. 261--266, 2015.

\bibitem{palatucci2009zero}
M.~Palatucci, D.~Pomerleau, G.~E. Hinton, and T.~M. Mitchell, ``Zero-shot
  learning with semantic output codes,'' in \emph{Proc. Advances in Neural Inf.
  Process. Syst.}, 2009, pp. 1410--1418.

\bibitem{karpathy2015deep}
A.~Karpathy and L.~Fei-Fei, ``Deep visual-semantic alignments for generating
  image descriptions,'' in \emph{Proc. IEEE Conf. Comput. Vision and Pattern
  Recognition}, 2015, pp. 3128--3137.

\bibitem{bergstra2013making}
J.~Bergstra, D.~Yamins, and D.~D. Cox, ``Making a science of model search:
  Hyperparameter optimization in hundreds of dimensions for vision
  architectures,'' \emph{J. Machine Learning Research}, vol.~28, 2013.

\bibitem{shahriari2016taking}
B.~Shahriari, K.~Swersky, Z.~Wang, R.~P. Adams, and N.~De~Freitas, ``Taking the
  human out of the loop: A review of {B}ayesian optimization,'' \emph{Proc.
  IEEE}, vol. 104, no.~1, pp. 148--175, 2016.

\bibitem{mikolov2013efficient}
T.~Mikolov, K.~Chen, G.~Corrado, and J.~Dean, ``Efficient estimation of word
  representations in vector space,'' \emph{arXiv preprint arXiv:1301.3781},
  2013.

\bibitem{pennington2014glove}
J.~Pennington, R.~Socher, and C.~Manning, ``Glo{V}e: Global vectors for word
  representation,'' in \emph{Proc. Conf. Empirical Methods in Natural Language
  Process.}, 2014, pp. 1532--1543.

\bibitem{mnih2013learning}
A.~Mnih and K.~Kavukcuoglu, ``Learning word embeddings efficiently with
  noise-contrastive estimation,'' in \emph{Proc. Advances in Neural Inf.
  Process. Syst.}, 2013, pp. 2265--2273.

\bibitem{lebret2013word}
R.~Lebret and R.~Collobert, ``Word emdeddings through {H}ellinger {PCA},''
  \emph{arXiv preprint arXiv:1312.5542}, 2013.

\bibitem{mikolov2010recurrent}
T.~Mikolov, M.~Karafi{\'a}t, L.~Burget, J.~{\v{C}}ernock{\`y}, and
  S.~Khudanpur, ``Recurrent neural network based language model,'' in
  \emph{Proc. Eleventh Annual Conf. Int. Speech Commun. Association}, 2010.

\bibitem{glove_pretrained}
``Glo{V}e pre-trained word vectors,''
  \url{https://nlp.stanford.edu/projects/glove/}, accessed: 02-10-2019.

\bibitem{word2vec_pretrained}
``word2vec pre-trained word vectors,''
  \url{https://code.google.com/archive/p/word2vec/}, accessed: 02-10-2019.

\bibitem{word2vec_pretrained_biomedical}
``word2vec pre-trained word vectors for biomedical applications,''
  \url{http://bio.nlplab.org}, accessed: 02-10-2019.

\bibitem{motwani2004survey}
M.~C. Motwani, M.~C. Gadiya, R.~C. Motwani, and F.~C. Harris, ``Survey of image
  denoising techniques,'' in \emph{Proc. Int. Pervasive Signal Process. Conf.
  Exhibition}, vol. 2004, 2004, pp. 27--30.

\bibitem{joshi2013survey}
S.~L. Joshi, R.~A. Vatti, and R.~V. Tornekar, ``A survey on {ECG} signal
  denoising techniques,'' in \emph{Proc. IEEE Int. Conf. Commun. Syst. Network
  Technologies}, 2013, pp. 60--64.

\bibitem{kandaswamy2005removal}
A.~Kandaswamy, V.~Krishnaveni, S.~Jayaraman, N.~Malmurugan, and K.~Ramadoss,
  ``Removal of ocular artifacts from {EEG} - {A} survey,'' \emph{Institution of
  Electron. Telecommunication Eng. J. Research}, vol.~51, no.~2, pp. 121--130,
  2005.

\bibitem{mohan2014survey}
J.~Mohan, V.~Krishnaveni, and Y.~Guo, ``A survey on the magnetic resonance
  image denoising methods,'' \emph{Biomedical Signal Process. Control}, vol.~9,
  pp. 56--69, 2014.

\bibitem{hodge2004survey}
V.~Hodge and J.~Austin, ``A survey of outlier detection methodologies,''
  \emph{Artificial Intelligence Review}, vol.~22, no.~2, pp. 85--126, 2004.

\bibitem{kelleher2015fundamentals}
J.~D. Kelleher, B.~Mac~Namee, and A.~D'Arcy, \emph{Fundamentals of Machine
  Learning for Predictive Data Analytics}.\hskip 1em plus 0.5em minus
  0.4em\relax MIT Press, 2015.

\bibitem{Dua:2017}
\BIBentryALTinterwordspacing
D.~Dua and E.~Karra~Taniskidou, ``{UCI} machine learning repository,'' 2017.
  [Online]. Available: \url{http://archive.ics.uci.edu/ml}
\BIBentrySTDinterwordspacing

\bibitem{bo_framework}
``Bayesian optimization framework,''
  \url{https://github.com/fmfn/BayesianOptimization}, accessed: 03-06-2019.

\bibitem{scikit-learn}
F.~Pedregosa, G.~Varoquaux, A.~Gramfort, V.~Michel, B.~Thirion, O.~Grisel,
  M.~Blondel, P.~Prettenhofer, R.~Weiss, V.~Dubourg, J.~Vanderplas, A.~Passos,
  D.~Cournapeau, M.~Brucher, M.~Perrot, and E.~Duchesnay, ``Scikit-learn:
  Machine learning in {P}ython,'' \emph{J. Machine Learning Research}, vol.~12,
  pp. 2825--2830, 2011.

\bibitem{van2001art}
D.~A. Van~Dyk and X.-L. Meng, ``The art of data augmentation,'' \emph{J.
  Computational and Graphical Statistics}, vol.~10, no.~1, pp. 1--50, 2001.

\bibitem{he2008adasyn}
H.~He, Y.~Bai, E.~A. Garcia, and S.~Li, ``{ADASYN}: Adaptive synthetic sampling
  approach for imbalanced learning,'' in \emph{Proc. IEEE Int. Joint Conf. on
  Neural Networks}, 2008, pp. 1322--1328.

\bibitem{freund1999short}
Y.~Freund, R.~Schapire, and N.~Abe, ``A short introduction to boosting,''
  \emph{Journal-Japanese Society For Artificial Intelligence}, vol.~14, no.
  771-780, p. 1612, 1999.

\bibitem{dietterich2002ensemble}
T.~G. Dietterich \emph{et~al.}, ``Ensemble learning,'' \emph{The Handbook of
  Brain Theory and Neural Networks}, vol.~2, pp. 110--125, 2002.

\bibitem{van2009dimensionality}
L.~Van Der~Maaten, E.~Postma, and J.~Van~den Herik, ``Dimensionality reduction:
  A comparative,'' \emph{J. Machine Learning Research}, vol.~10, no. 66-71,
  p.~13, 2009.

\bibitem{alom2019state}
M.~Z. Alom, T.~M. Taha, C.~Yakopcic, S.~Westberg, P.~Sidike, M.~S. Nasrin,
  M.~Hasan, B.~C. Van~Essen, A.~A. Awwal, and V.~K. Asari, ``A state-of-the-art
  survey on deep learning theory and architectures,'' \emph{Electronics},
  vol.~8, no.~3, p. 292, 2019.

\bibitem{liu2018task}
Q.~Liu, H.~Huang, Y.~Gao, X.~Wei, Y.~Tian, and L.~Liu, ``Task-oriented word
  embedding for text classification,'' in \emph{Proc. Int. Conf. Computational
  Linguistics}, 2018, pp. 2023--2032.

\bibitem{kusner2015word}
M.~Kusner, Y.~Sun, N.~Kolkin, and K.~Weinberger, ``From word embeddings to
  document distances,'' in \emph{Proc. Int. Conf. Machine Learning}, 2015, pp.
  957--966.

\bibitem{bordes2014open}
A.~Bordes, J.~Weston, and N.~Usunier, ``Open question answering with weakly
  supervised embedding models,'' in \emph{Proc. Joint European Conf. Machine
  Learning and Knowledge Discovery in Databases}.\hskip 1em plus 0.5em minus
  0.4em\relax Springer, 2014, pp. 165--180.

\bibitem{bengio2014word}
S.~Bengio and G.~Heigold, ``Word embeddings for speech recognition,'' in
  \emph{Proc. Fifteenth Annual Conf. Int. Speech Commun. Association}, 2014,
  pp. 1053--1057.

\bibitem{wang2016cnn}
J.~Wang, Y.~Yang, J.~Mao, Z.~Huang, C.~Huang, and W.~Xu, ``{CNN-RNN}: A unified
  framework for multi-label image classification,'' in \emph{Proc. IEEE
  Conference on Computer Vision and Pattern Recognition}, 2016, pp. 2285--2294.

\bibitem{socher2013zero}
R.~Socher, M.~Ganjoo, C.~D. Manning, and A.~Ng, ``Zero-shot learning through
  cross-modal transfer,'' in \emph{Proc. Advances in Neural Information
  Processing Syst.}, 2013, pp. 935--943.

\bibitem{liu2015learning}
P.~Liu, X.~Qiu, and X.~Huang, ``Learning context-sensitive word embeddings with
  neural tensor skip-gram model,'' in \emph{Proc. Twenty-Fourth Int. Joint
  Conference on Artificial Intelligence}, 2015.

\bibitem{liu2015topical}
Y.~Liu, Z.~Liu, T.-S. Chua, and M.~Sun, ``Topical word embeddings,'' in
  \emph{Proc. AAAI Conf. Artificial Intelligence}, 2015.

\bibitem{nguyen2017overview}
D.~Q. Nguyen, ``An overview of embedding models of entities and relationships
  for knowledge base completion,'' \emph{arXiv preprint arXiv:1703.08098},
  2017.

\bibitem{lee2017muse}
G.-H. Lee and Y.-N. Chen, ``{MUSE}: Modularizing unsupervised sense
  embeddings,'' \emph{arXiv preprint arXiv:1704.04601}, 2017.

\bibitem{jasper1971acetylcholine}
H.~H. Jasper and J.~Tessier, ``Acetylcholine liberation from cerebral cortex
  during paradoxical (rem) sleep,'' \emph{Science}, vol. 172, no. 3983, pp.
  601--602, 1971.

\bibitem{greenberg1972effect}
R.~Greenberg, R.~Pillard, and C.~Pearlman, ``The effect of dream (stage {REM})
  deprivation on adaptation to stress.'' \emph{Psychosomatic Medicine}, 1972.

\bibitem{resnik2010evaluation}
P.~Resnik and J.~Lin, ``Evaluation of {NLP} systems,'' \emph{The Handbook of
  Computational Linguistics and Natural Language Processing}, vol.~57, pp.
  271--295, 2010.

\bibitem{zhai2016intrinsic}
M.~Zhai, J.~Tan, and J.~D. Choi, ``Intrinsic and extrinsic evaluations of word
  embeddings.'' in \emph{Proc. Association for the Advancement of Artificial
  Intell.}, 2016, pp. 4282--4283.

\end{thebibliography}

\end{document}